\documentclass[10pt,twocolumn,letterpaper]{article}
\usepackage[accsupp]{axessibility} 
\usepackage{iccv}
\usepackage{times}
\usepackage{epsfig}
\usepackage{graphicx}
\usepackage{amsmath}
\usepackage{bm}
\usepackage{amssymb}
\usepackage{ulem}
\usepackage{enumitem}
\usepackage{xcolor}
\usepackage{blindtext}%
\usepackage{ulem}
\definecolor{green}{RGB}{102,252,102}
\usepackage{bm}
\usepackage{soul}
\usepackage[accsupp]{axessibility}  % Improves PDF readability for those with disabilities.

\usepackage{hyperref}
\hypersetup{pagebackref=true,breaklinks=true,letterpaper=true,colorlinks,bookmarks=false}

\iccvfinalcopy % *** Uncomment this line for the final submission

\def\iccvPaperID{7119} % *** Enter the ICCV Paper ID here

\begin{document}
% \begin{titlepage}
%%%%%%%%% TITLE
\title{Exploiting Explanations for Model Inversion Attacks}

\author{Xuejun Zhao
\qquad
Wencan Zhang
\qquad
Xiaokui Xiao
\qquad
Brian Lim\thanks{Corresponding author.}\\
National University of Singapore, Singapore\\
{\tt\small \{xuejunzhao, wencanz\}@u.nus.edu, xkxiao@nus.edu.sg, brianlim@comp.nus.edu.sg}
% \and
}

\maketitle
% \thispagestyle{empty}
% \end{titlepage}
% \newpage
\thispagestyle{empty}
% \thispagestyle{plain}
% \pagestyle{plain}

%%%%%%%%% ABSTRACT
\begin{abstract}
% 4000 chars limit
   %Model inversion attack is scary as attacker can invert training image from prediction result. When meeting explanation from XAI API, model inversion attack can be strong to invert recognizable face image. In this work, several explanations are involved to present how strong model inversion attack is. Moreover, the performance has been proved to be scarier than obfuscation. 
   
   The successful deployment of artificial intelligence (AI) in many domains from healthcare to hiring requires their responsible use, particularly in model explanations and privacy. 
   Explainable artificial intelligence (XAI) provides more information to help users to understand model decisions, yet this additional knowledge exposes additional risks for privacy attacks. Hence, providing explanation harms privacy. 
   We study this risk for image-based model inversion attacks and identified several attack architectures with increasing performance to reconstruct private image data from model explanations. 
   We have developed several multi-modal transposed CNN architectures that achieve significantly higher inversion performance than using the target model prediction only. These XAI-aware inversion models were designed to exploit the spatial knowledge in image explanations. 
   To understand which explanations have higher privacy risk, we analyzed how various explanation types and factors influence inversion performance. 
   In spite of some models not providing explanations, we further demonstrate increased inversion performance even for non-explainable target models by exploiting explanations of surrogate models through attention transfer. This method first inverts an explanation from the target prediction, then reconstructs the target image.
   These threats highlight the urgent and significant privacy risks of explanations and calls attention for new privacy preservation techniques that balance the dual-requirement for AI explainability and privacy.
\vspace{-0.7cm}
\end{abstract}

%%%%%%%%% BODY TEXT
\section{Introduction}

\begin{figure}[t]
    \centering
    \includegraphics[width=8.3cm]{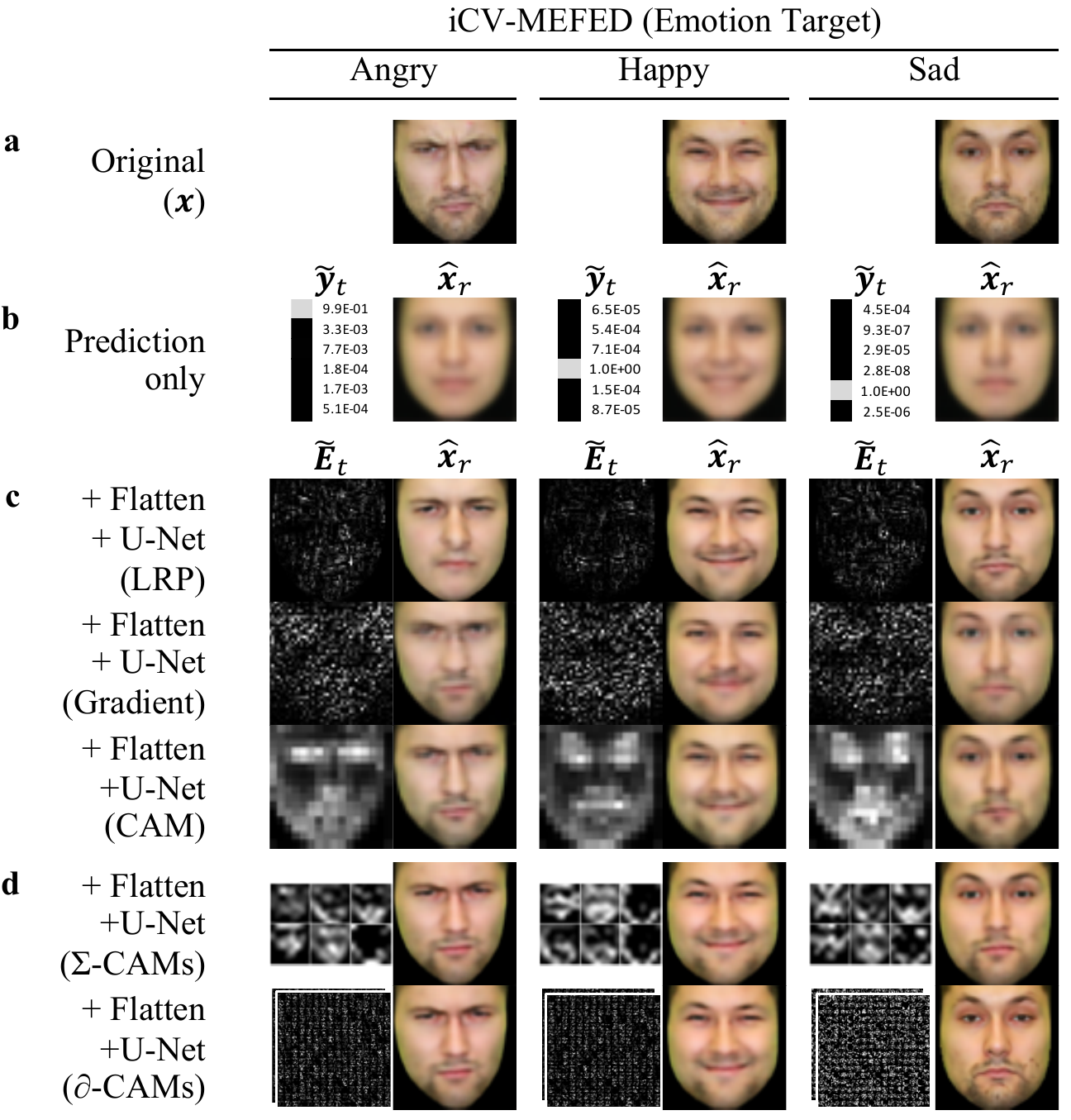}
    \caption{
    Demonstration of image reconstruction from XAI-aware inversion attack with emotion prediction as target task, and face reconstruction as attack task. Three emotions from a single identity shown from the iCV-MEFED dataset \cite{loob2017dominant}. Reconstructions shown with corresponding inputs (target prediction $\tilde{\bm{y}}_t$ and explanations $\tilde{\bm{E}}_t$ as LRP \cite{bach2015pixel},Gradients \cite{simonyan2013deep} or Grad-CAM \cite{selvaraju2017grad} saliency maps). Towards original images (a), reconstructions from Prediction only (b) are poor and similar across different faces, and is significantly improved when exploiting single (c) and multiple (d) explanations. 
    Demonstration reconstructions for other explanations and baseline inversion models are shown in supplementary materials.
    }
    \label{fig:demoEmotion}
    \vspace{-.5cm}
\end{figure}

The recent success of artificial intelligence (AI) is driving its application in many domains from healthcare to hiring. 
With the increasing regulatory requirements for responsible AI, these applications will need to support both explainability to justify important decisions and privacy to protect personal information \cite{regulation2016regulation}.
Indeed, many AI applications have this dual-requirement, such as driver drowsiness detection from faces \cite{gao2014detecting,reddy2017real} that can be used to limit the jobs of tired drivers, and classroom facial engagement prediction \cite{bosch2016detecting,whitehill2014faces} for student grading or teacher performance evaluation. Users need explanations to dispute or rectify unfavorable model predictions, and need privacy to preserve anonymity or reputation (e.g., of embarrassing appearances). Troublingly, using state-of-the-art model inversion attacks \cite{fredrikson2015model, yang2019neural, zhang2020secret}, attackers can reconstruct sensitive information (e.g., faces) merely from model predictions. We hypothesize that since explanations provide more information, they can be used by attackers for more effective reconstructions, and further harm privacy
% . Clearly there is a conflict between these two requirements 
--- providing explanations harms privacy.

Explainable artificial intelligence (XAI) provides information to help users to debug models \cite{kulesza2009fixing}, improve decision making \cite{wang2019designing}, and improve and moderate trust in automation \cite{dzindolet2003role}. 
% XAI techniques include explaining with attributions \cite{datta2016algorithmic,lundberg2017unified,ribeiro2016should}, rules and trees \cite{letham2015interpretable,ribeiro2018anchors}, influential and counterfactual examples \cite{kim2016examples, koh2017understanding,wachter2017counterfactual}, partial dependence plots and generalized additive models \cite{caruana2015intelligible,krause2016interacting}, etc. 
Specific to image-based deep learning, XAI techniques include saliency maps \cite{chattopadhay2018grad,ramaswamy2020ablation,simonyan2013deep,selvaraju2017grad,zhou2016learning}, feature visualization \cite{bau2017network,hohman2019s,olah2017feature}, and activations of neurons \cite{kahng2017cti} and concept vectors \cite{kim2018interpretability,koh2020concept}. 
These explanations provide users with deeper insights into model reasoning and about the data, but they can contain sensitive knowledge that can be exploited for privacy attacks, especially with recent machine learning (ML)-based attacks \cite{shokri2017membership,tramer2016stealing,fredrikson2015model}.
%
% While conventional privacy attacks involve linking structured data from different sources to de-anonymize users, ML-based attacks can learn complex patterns to leak private information from unstructured data, such as images. 
% Several privacy attacks have been proposed for membership inference to identify if a user's data was in a dataset \cite{shokri2019privacy}, model extraction to steal confidential data of model weights \cite{milli2019model}, attribute inference for data imputation \cite{tramer2016stealing}, and model inversion to reconstruct training or test data \cite{fredrikson2015model,yang2019neural,zhang2020secret}.
We focus on model inversion attacks that can reconstruct data from model predictions \cite{fredrikson2015model,he2019model,yang2019neural,zhang2020secret}.
These methods typically require some privileged white-box access (to model parameters) or background knowledge (such as blurred images) for effective attack, but this is unrealistic or uncommon in deployed settings. Instead, because of the requirement for explainability, model explanations will be more readily available and pose a more ubiquitous threat.

Recent work has begun to study the privacy threats of model explanations, but their attack goals differ and their use of explanations remains under-studied. Instead of attacking training data \cite{shokri2017membership} or model confidentiality \cite{aivodji2020model,milli2019model}, we focus on attacking the privacy of test instances. 
This will compromise the trust of active users of the model. 
We propose attack methods based on the structure and generation technique of explanations, focusing on saliency maps that are highly popular for image-based AI. 
% Having identified that providing explanations can compromise privacy, developers may be dis-incentivised provide model explanations. This can limit the usefulness and trust in AI and violate regulations for responsible AI. 
% However, we further show that leveraging on XAI techniques with attention transfer, inversion attacks can be made stronger by exploiting an explainable surrogate model instead. 
Furthermore, we propose a method of leveraging on XAI techniques with attention transfer to perform inversion attacks on models that do not share explanations.
We found that this performance is close to inverting an explainable target model. Hence, even non-explainable target models are at increased risk of XAI-aware inversion attacks. 

\textbf{Our contributions are:}
\textbf{1)} We determine the privacy threat of model explanations for model inversion attacks. We achieved significantly higher inversion performance by proposing several architectures for XAI-aware model inversion, through careful adapting of multi-modal, spatially-aware transposed CNNs. This highlights that the present privacy risk of providing explanations is significant (see Figure \ref{fig:demoEmotion}).
\textbf{2)} We propose an XAI-aware attack on non-explainable target models that can achieve improved inversion performance using an explanation inversion model with surrogate explanation attention transfer. This does not need additional data leak or sharing from the target model, and demonstrates a significantly increased risk due to explanations in surrogate models instead of the target model. 
\textbf{3)} We identify the privacy risk for different explanation types (Gradients \cite{simonyan2013deep}, Grad-CAM \cite{selvaraju2017grad}, and layer-wise relevance propagation (LRP) \cite{bach2015pixel}) and analyze how various factors influence attack performance.
% \begin{comment}
% \begin{itemize}[noitemsep,topsep=0pt]
%     \item We determine the privacy threat of model explanations for model inversion attacks. We achieved significantly higher inversion performance by proposing several architectures for XAI-aware model inversion, through careful adapting of multi-modal, spatially-aware transposed CNNs. This highlights that the present privacy risk of providing explanations is significant.
%     \item We propose an XAI-aware attack on non-explainable target models that can achieve improved inversion performance using an explanation inversion model with surrogate explanation attention transfer. This does not need additional data leak or sharing from the target model, and demonstrates a significantly increased risk due to explanations in surrogate models instead of the target model. 
%     \item We identify the privacy risk for different explanation types (e.g., gradients \cite{lecun1998gradient}, Grad-CAM \cite{selvaraju2017grad}) and analyze how various factors influence attack performance.
% \end{itemize}
% \end{comment}

In this work, we provide the first study into the privacy risk of explanations for inversion attack, and defer their defense to future work. This highlights the urgency to develop privacy defense techniques and models that are co-optimized for the dual-requirements of explainability and privacy to achieve the objectives of Responsible AI.

\section{Related Works}

% - introduce describe usefulness of explanations
% - introduce ML-based privacy attacks
% - focus on model inversion attacks that can reconstruct private images.
% - discuss early work on privacy attacks with model explanations, but still open regarding model inversion attack.

Our work relates to two research areas --- explainable AI and machine learning-based privacy --- which we overview separately, and discuss nascent work intersecting the two.

% \textbf{Explainable AI.}
%Despite the accuracy of deep learning models for many real-world prediction tasks, their complexity and unintelligibility hampers their adoption. This has driven the burgeoning field of explainable AI (XAI) \cite{gunning2017explainable} to help users to debug models \cite{kulesza2009fixing}, improve decision making \cite{wang2019designing}, and improve and moderate trust in automation \cite{dzindolet2003role}. 

% \textbf{Image CNN model explanations.} 
\textbf{Explainable AI for image-based CNN models.} 
 While there are many XAI techniques 
% \cite{caruana2015intelligible,datta2016algorithmic,kim2016examples,krause2016interacting,koh2017understanding,letham2015interpretable,lundberg2017unified,ribeiro2016should,ribeiro2018anchors,wachter2017counterfactual}, 
(see review \cite{biran2017explanation,gilpin2018explaining}),
we focus on explanations of image-based convolutional networks (CNN) 
% \cite{bau2017network,hohman2019s,kahng2017cti,kim2018interpretability,koh2020concept,olah2017feature}, 
(see review \cite{zhang2018visual})
and specifically saliency map explanations that highlight important pixels for each prediction \cite{chattopadhay2018grad,ramaswamy2020ablation,simonyan2013deep,selvaraju2017grad,zhou2016learning}. 
Gradients \cite{simonyan2013deep} of the model prediction with respect to input features, i.e., pixels in an image, describe the model sensitivity. Though easy to calculate, this suffers from several issues, which are addressed in extensions, such as robustness (SmoothGrad \cite{smilkov2017smoothgrad}), misleading saturated gradients (DeepLIFT \cite{shrikumar2016not}), and implementation invariance (Integrated Gradients \cite{sundararajan2017axiomatic}).
For this work, we study the exploitation of the general gradient technique \cite{simonyan2013deep}, and expect that our findings will generalize to the extensions.
Strictly, these techniques produce sensitivity maps and describe how much the prediction may change if the pixels are altered. Alternatively, activation map explanations show how influential each pixel for the prediction. One simple approach is $\text{Gradient} \odot \text{Input}$ \cite{shrikumar2016not} which is the element-wise product of gradient and image to approximate the model activation based on the image, but this highlights fine-grained details that may be hard to read. 

Class activation maps \cite{chattopadhay2018grad,ramaswamy2020ablation,selvaraju2017grad,zhou2016learning} provide class-specific coarse-grained explanations based on activation maps of convolution filters. While CAM \cite{zhou2016learning} required model retraining, Grad-CAM \cite{selvaraju2017grad} could be read from any CNN without retraining. Extensions improved explanations for robustness (Ablation-CAM \cite{ramaswamy2020ablation}) and to handle multiple objects (Grad-CAM++ \cite{chattopadhay2018grad}). 
For this work, we evaluated with the original Grad-CAM \cite{selvaraju2017grad} approach, and expect that our findings will generalize to its extensions.
While the aforementioned techniques are model-specific and require white box access to model parameters, model-agnostic approaches can apply generally to any deep neural network, such as perturbation-based sensitivity analysis \cite{zeiler2014visualizing}, and layer-wise relevance propagation (LRP) \cite{bach2015pixel}. However, as proxy explanations, these techniques are less faithful to the model behavior, and we defer their study.
%We defer the study of this XAI technique, but it presents a good opportunity for tunable explanations to trade-off faithfulness for privacy.

Other than helping user understanding, saliency map explanations have also been used to improve model training through transfer learning from a better model with student-teacher networks \cite{komodakis2017paying}. Attention can also be indirectly transferred by maximizing the loss between predicting on an image and its variant ablated with its saliency mask (GAIN \cite{li2018tell}). Similarly, to attack non-explainable target models, we exploit attention transfer to minimize the intermediate explanation in the attack model and the explanation of the surrogate target model.

\textbf{Machine learning privacy and model inversion attacks.}
% Just as how AI can improve model decision quality, it can also be used for malicious tasks. 
Recent research on machine learning privacy has identified sophisticated attacks, such as
% to infer private information from unstructured data. 
model extraction attacks \cite{tramer2016stealing,salem2020updates} to reconstruct parameters of proprietary models,
% and this has been been used to infer knowledge about the training data \cite{song2017machine}. 
membership inference attacks \cite{shokri2017membership} to identify if users were part of a training dataset (e.g., of cancer patients \cite{jia2019memguard}), 
attribution inference attacks \cite{fredrikson2014privacy} to impute omitted or obfuscated data to recover the sensitive information, 
and
model inversion attacks \cite{fredrikson2015model,yang2019neural,zhang2020secret} to infer the original data from the target prediction (e.g., reconstructing a face from an emotion prediction). 
In this work, we focus on the latter attack, which is relevant to image-based machine learning, where private input images can be recovered, thus de-anonymizing users or revealing sensitive details. 
%
% \textbf{Model inversion attacks.}
While the original model inversion attack has limited performance \cite{fredrikson2015model}, inversion performance is notably improved by leveraging deep architectures \cite{he2019model,yang2019neural} (specifically, transposed CNNs \cite{dosovitskiy2016inverting}). Further improvements exploit auxiliary knowledge, such as white-box model access \cite{zhang2020secret}, feature embeddings \cite{he2019model}, blurred images \cite{zhang2020secret}, 
or the joint probability distribution of features and labels \cite{yeom2018privacy}, 
but such information is not readily available in practice. However, the need for explanations will increase their ubiquity and this will poses increased privacy risk. 
% Specifically, we hypothesize that while providing model explanations aids user understanding, it leaks sensitive information; e.g., a model inversion attack using both the target model prediction and saliency map explanation can increase the image reconstruction accuracy compared to just using the prediction.

\textbf{Privacy risk of model explanations.} 
Membership inference attacks can be improved by concatenating gradient explanations with model predictions \cite{shokri2019privacy}.
Model extraction attacks can be improved by regularizing the reconstructed model with gradient explanations \cite{milli2019model}, and
training a surrogate model from counterfactual explanation examples \cite{aivodji2020model}.
However, exploiting explanations for model inversion attacks remains unexplored; this conceals the privacy risk on user data at prediction time, i.e., of active users. In this work, we investigate how to exploit different saliency map explanations types to attack explainable and non-explainable models for inversion attacks.

\section{XAI-Aware Model Inversion Attack}
\begin{figure}[t]
    \centering
    \hspace*{-0.5cm}
    \includegraphics[width=8.2cm]{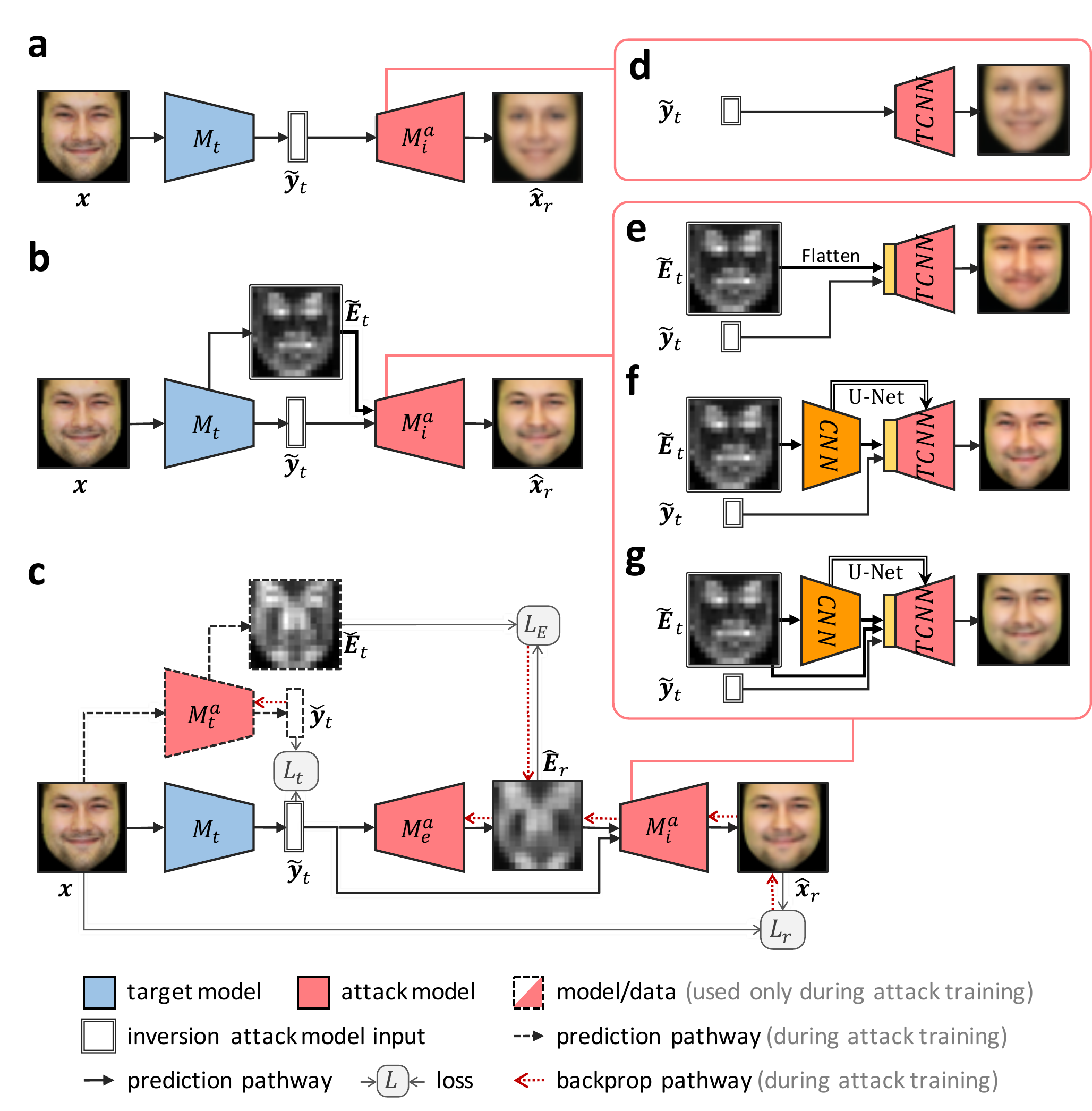}
    \caption{
    Architectures of inversion attack models. 
    a) Baseline threat model with target CNN model $M_t$ to predict emotion $\tilde{\bm{y}}_t$ from face $\bm{x}$, and inversion attack model to reconstruct face $\hat{\bm{x}}_r$ from emotion. Emotion prediction confidences are input to a transposed CNN (TCNN) for inversion attack (d). 
    b) Threat model with explainable target model that also provides explanation $\tilde{\bm{E}}_t$ of the target prediction, and XAI-aware multi-modal inversion attack model that inputs $\tilde{\bm{E}}_t$ via different input architectures: e) Flattened $\tilde{\bm{E}}_t$ concatenated with $\tilde{\bm{y}}_t$, f) U-Net for dimensionality reduction and spatial knowledge, g) combined Flatten and U-Net. Additional architectures shown in supplementary materials. 
    c) Threat model with non-explainable target model, and inversion attack model that predicts a reconstructed surrogate explanation $\hat{\bm{E}}_r$ from target prediction $\tilde{\bm{y}}_t$ and uses $\hat{\bm{E}}_r$ for multi-modal image inversion (e-g). 
    %Explanation inversion is trained with attention transfer from surrogate target model $M_t^a$.
    % Target models are colored blue, models trained by attacker in red. Double solid lines represent inversion model input; solid lines represent pathway for prediction; dashed lines represent models, data, predictions, or pathways used only during training and not at prediction; grey solid lines indicate loss comparisons; dotted red lines indicate backpropagation pathway during training, showing which models get updated.
    }
    \label{fig:models}
    \vspace{-0.5cm}
\end{figure}

We describe the general approach for model inversion and describe how to exploit explanations for more aggressive attacks through various architectures (see Figure \ref{fig:models}).

\subsection{Threat Model}
Consider a target model ($M_t$) that has been trained and is deployed for use through an application programming interface (API). It takes private data $\bm{x}\in X_p$ (e.g., face image) provided by a user to produce a \textit{target prediction} $\tilde{\bm{y}_t}$ (e.g., emotion) as a confidence probability vector. It is trained with cross-entropy loss $L_t(\bm{y}_t,\tilde{\bm{y}})$, where $\bm{y}_t$ is the ground truth label for the target task.
To improve user trust and verification, the target model also provides a \textit{target explanation} $\tilde{\bm{E}}_t$ for each prediction (e.g., saliency map \cite{selvaraju2017grad}). 
Consider a model inversion attack at run time, where the attacker obtains access to each tuple of the target prediction vector and explanation tensor (e.g., due to breached storage, intercepted in transit, shared via social media). We assume that the attacker only has access to the breached data, an independent dataset $\bm{x}\in X_a$, and the target model via its API (i.e., black-box access, unlike white-box access in \cite{fredrikson2015model, zhang2020secret}). Furthermore, unlike \cite{yang2019neural, zhang2020secret}, we do not require privileged background knowledge (e.g., blurred images).
The attack goal is to train an inversion attack model $M_i^a$ to reconstruct the original image $\bm{x}$ from the model outputs $(\tilde{\bm{y}}_t,\tilde{\bm{E}}_t)$ such that sensitive information can be predicted from the reconstructed image $\hat{\bm{x}}_r$. Using facial emotion recognition as the running example for the target model, the attack goal is to reconstruct the face image and re-identify the user. This poses problems of consent and identity theft. 
In our case, the target and attack tasks are not identical; this differs from prior works that have identity prediction for both target and attack tasks \cite{fredrikson2015model,zhang2020secret,yang2019neural}. Figure \ref{fig:models}a illustrates the baseline model inversion attack using only target model predictions. We will next describe more serious attacks by exploiting target model explanations.

\subsection{Model Inversion with Target Explanations}
To invert the target model $M_t$, \cite{yang2019neural}, we trained the inversion attack model $M_i^a$ as a Transposed CNN (TCNN) \cite{dumoulin2016guide} to predict a 2D image $\hat{\bm{x}}_r$ from the 1D target prediction vector $\tilde{\bm{y}}_t$ as input to the attack model (see Figure \ref{fig:models}a). $M_i^a$ is trained with MSE loss for the image reconstruction $L_r(\bm{x},\hat{\bm{x}}_r)$. We illustrate the convolutional layers for upsampling in the TCNN in the supplementary materials. 
We consider saliency map explanations (e.g., \cite{simonyan2013deep, selvaraju2017grad}) and extend the TCNN model to be multi-modal to include the 2D target explanation as a second input (Figure \ref{fig:models}b). A simple approach is to flatten the explanation into a 1D vector, concatenate with the prediction vector and input to the TCNN (Figure \ref{fig:models}e). 
However, this neglects the spatial information of the salient pixels. A simple way to leverage the 2D information is with a CNN architecture that uses convolutional kernels to interpret 2D patterns into a 1D feature embedding vector which is put into the TCNN (see supplementary materials). This CNN-TCNN architecture is similar to the generator proposed in \cite{zhang2020secret}. 
Inspired by CNN encoder-decoder networks \cite{rahim2018end} and image super-resolution techniques \cite{zhang2018residual}, we propose to use a U-Net architecture \cite{ronneberger2015u} in the inversion model to improve its reconstruction performance (Figure \ref{fig:models}f). \
%textcolor{green}{We are the 1st to use U-Net for inversion attacks and our asymmetric architecture (fewer layers to encode CAM, more to decode image) is unique.}
We hypothesize that this will have more accurate reconstruction because of the retention in spatial information that is transmitted through the bypass connectors from the convolutional layers. 
We further propose a hybrid model by combining the Flatten and U-Net input architectures as the Flatten+U-Net inversion model (Figure \ref{fig:models}g); the 1D flattened explanation and prediction vectors are concatenated with the latent layer of the U-Net architecture. The proposed architectures demonstrate various methods to obtain sensitive information for privacy attack; we heretofore call them XAI Input Methods. The training loss function regards image reconstruction task:
\vspace{-0.2cm}
\begin{equation}
    \label{eq:reconstructionLoss}
    L_r=\sum_x (M_i^a(M_t(\bm{x})) - \bm{x})^2
    \vspace{-0.2cm}
\end{equation}
% consider how other papers, e.g., Secret Revealer writes their math
where $\bm{x}$ is the private input image, $M_t(\bm{x})=\tilde{\bm{y}_t}$ is the target model prediction, and $M_i^a(M_t(\bm{x}))=\hat{\bm{x}}_r$ is the reconstructed image from the inversion attack.

Later, we describe our experiments on how some architectures capture more knowledge from the explanations and interpret why. The proposed architectures can be used for any 2D explanations of the CNN target model, such as gradients \cite{simonyan2013deep}, class activation maps (CAM) \cite{zhou2016learning}, LRP \cite{bach2015pixel}, and 2D auxiliary data, such as blurred variants of the input images $x_b$. We further conducted experiments on different Explanation Types, described later.

\subsection{Model Inversion with Multiple Explanations}

While many explanations explain \textit{why} a model predicted a class $c \in C$, it is also important to explain \textit{why not} an alternative class $c' \neq c$, i.e., to provide contrastive explanations \cite{lim2009and,miller2019explanation}. 
To support this, some explanation techniques, such as Grad-CAM \cite{selvaraju2017grad}, can provide class-specific explanations depending on user query. However, this further raises the privacy risk, since more information is provided by the additional explanations. We exploit these Alternative CAMs ($\Sigma$-CAM) by concatenating explanations for $|C|$ classes into a 3D tensor and training the inversion models on this instead of the 2D matrix of a single explanation. Supplementary Figure \ref{fig:xaiInputs_high_level}f for detailed architecture.

To gain a deeper insight into a model decision, users may be interested in understanding which specific neurons of a CNN were activated and how. In particular, although Grad-CAM presents one saliency map per class $c$, this is composed as a weighted sum of activation maps from the last convolutional layer in the CNN, i.e., the saliency map explanation is $E^c = \text{ReLU}{\left(\sum_{k}{\alpha_k^c A^k}\right)}$, where $A^k$ is the $k$th activation map ($k \in K$), and $\alpha_k^c=\frac{1}{HW}\sum_{i,j}^{H,W} \frac{\partial y^c}{\partial A_{ij}^k}$ is the gradient-based importance weight. Hence, the Grad-CAM explanation is composed of $|K|$ partial, constituent CAMs ($\partial$-CAM). Combined with explanation techniques to understand the role of each neuron (e.g., \cite{bau2017network,olah2017feature}), these $\partial$-CAMs can provide rich insights for developers. However, since $|K|>\!\!>1$ is large, this provides a multitude of explanations that further leak privacy. Similar to Alternative CAMs, we exploit these Constituent CAMs $\partial$-CAM by concatenating them into a 3D tensor as input to the inversion model.

%\subsection{Model Inversion without Target Explanations}
\subsection{Model Inversion with Surrogate Explanations}
While the XAI Input methods can increase privacy risk, we further hypothesize that XAI techniques can be exploited for inversion attacks even for non-explainable target models (i.e., no target explanation). We propose an architecture that predicts the target explanation and exploits that explanation to invert the original target data. 
Figure \ref{fig:models}c illustrates the architecture for this attack. First, we train an explainable surrogate target model $M_t^a$ on the attacker's dataset to generate surrogate explanation $\check{\bm{E}}_t$. 
% \textcolor{green}{Attacker dataset $X_a$ utilizes IID and less similar out-of-distribution (OOD) data to target dataset $X$.} 
%As $X_a$ is similar to the target model dataset $X$, then $M_t^a \approx M_t$, and the surrogate explanation will be similar to the target explanation, i.e., $\check{\bm{E}}_t \approx \tilde{\bm{E}}_t$. 
Then, as $X_a \to X$, then $M^a_t \to M_t$ and $\check{E} \to \tilde{E}$.
 
However, $\check{\bm{E}}_t$ is only available during training not when predicting. 
Hence, we train an explanation inversion model $M_e^a$ to reconstruct $\check{\bm{E}}_t$ as $\hat{\bm{E}}_r$ from the target prediction $\tilde{\bm{y}}_t$ by minimizing the surrogate explanation loss function:
\vspace{-0.2cm}
\begin{equation}
    \label{eq:explanationLoss}
    L_E=\sum_x (M_e^a(M_t(\bm{x})) - E(M_t^a(\bm{x})))^2
    \vspace{-0.2cm}
\end{equation}
where $E(M)$ is the explanation of the model $M$, $M_t^a(\bm{x})=\check{\bm{y}_t}$ is the surrogate target prediction, $E(M_t^a(\bm{x}))=\check{\bm{E}_t}$ is the surrogate explanation, and $M_e^a(M_t(\bm{x}))=\hat{\bm{E}}_r$ is the reconstructed surrogate explanation.
This reconstructed explanation is available at prediction time. 
Finally, we input $\hat{\bm{E}}_r$ into the image inversion model $M_i^a$ to complete the model inversion attack. Since $\hat{\bm{E}}_r$ is the same format as $\tilde{\bm{E}}_t$, we can apply any of the aforementioned XAI Input Methods. 
Overall, the proposed architecture can also be described as model inversion attack with attention transfer, where explanations are transferred from surrogate target model $M_t^a$ into the intermediate layer between $M_e^a$ and $M_i^a$. This guides and constrains the inversion model to learn activations that $M^a_t$ finds relevant for $\check{y}$ as a proxy for $\hat{y}$.
The training loss involves two tasks: image reconstruction (Eq. \ref{eq:reconstructionLoss}) and explanation reconstruction (Eq. \ref{eq:explanationLoss}).

\section{Experiments}

\begin{figure*}[t]
    \centering
    \includegraphics[width=17.4cm]{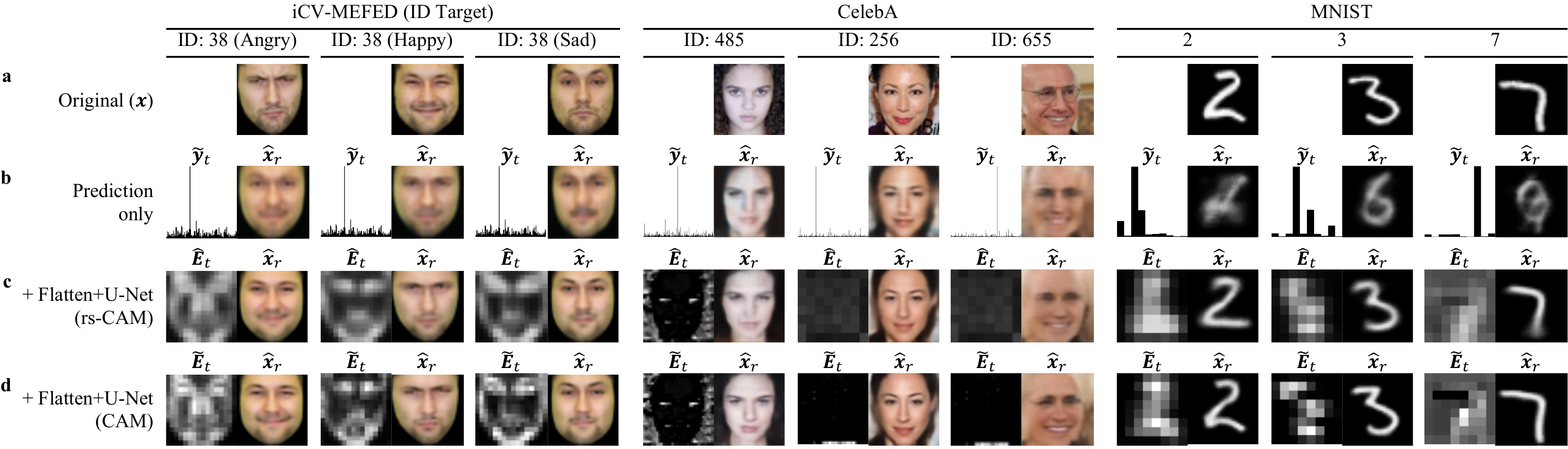}
    \caption{
    Demonstration of image reconstruction from XAI-aware inversion attack for different datasets with same target and attack tasks for each case (iCV-MEFED \cite{loob2017dominant} and CelebA \cite{liu2018large}: identification, MNIST \cite{lecun1998gradient}: handwriting digit recognition). Same format as Figure \ref{fig:demoEmotion}.
    % CelebA faces: Madison Pettis (?), Ann Curry, Larry David
    }
    \label{fig:demoId}
    \vspace{-0.3cm}
\end{figure*}

We conducted experiments to perform an ablation analysis of our model inversion architectures and to compare with existing work. We evaluated on multiple datasets for different use cases and evaluated image reconstruction quality and accuracy to classify sensitive information from reconstructed images. Figure \ref{fig:demoId} demonstrates increased inversion risk due to explanations and XAI-aware attention transfer.

\subsection{Experiment Setup}

\textbf{Use Cases and Datasets.} 
We evaluated on datasets representing three use cases with the simultaneous need for model explanations and privacy. 
1) iCV-MEFED face expressions data \cite{loob2017dominant} with 6 emotions and 115 identities (10 omitted due to confidentiality) over 24,120 instances.
We evaluated a first use case where an explainable target model predicts emotion from faces (e.g., for classroom engagement monitoring \cite{bosch2016detecting,whitehill2014faces}, driver drowsiness detection \cite{gao2014detecting,reddy2017real}); and an attacker executes a data breach to obtain the prediction and explanation data, and performs a model inversion attack \cite{fredrikson2015model, zhang2020secret} to reconstruct the face from the emotion prediction to re-identify the user (e.g., leak embarrassing or compromising faces). Users can interpret explanations to validate or dispute the emotion predictions (e.g., to defend their alertness), but they are at increasing risk of being de-anonymized. 
We evaluated a second use case of faces identity recognition for biometrics as target task with explainability to help human inspection (e.g., security access and passport border control) and privacy to mitigate identity theft. 
2) CelebFaces Attributes Dataset (CelebA) \cite{liu2018large} with a balanced subset of 1000 identities over 30,000 instances. This enables testing more realistic images (Internet scraped vs. lab captured) and more varied labels (1000 vs 115). For simplicity, all images were cropped to tighten on the face and exclude backgrounds.
% While the inversions will reconstruct greyscale images, image colorization methods can be employed to recolor them or reconstruct with accurate color (e.g., \cite{zhang2016colorful, iizuka2016let}).
3) MNIST handwritten digits \cite{lecun1998gradient} with 10 labels and 70,000 instances to test another use case of explainable handwriting biometrics. 
To fit our models, we resized iCV-MEFED to 128$\times$128, CelebA to 256$\times$256, and MNIST to 32$\times$32 pixels.

% z; maybe use different format
\textbf{Protocol.} 
We split each dataset into two disjoint sets: 50\% as target dataset to train the target model, 50\% as attack dataset to train and test the attack models. We tested the target model with the attack dataset. For the attack dataset, we employ an 80/20 ratio for the train/test split.
% iCV target training: 12060, target testing: 12060, attack training: 9648, attack testing:2412, total: 24120
% CelebA target training: 10576, target testing: 10576, attack training: 8461, attack testing:2115, total: 21152
% MNIST target training: 35000, target testing: 35000, attack training: 28000, attack testing:7000, total: 70000

% The data splitting is required for both target modeling phrase and attack modeling phrase. First, we split each dataset into two disjoint parts: one part used as the private dataset to train the target network and the other as a public dataset to test. Normally, we split the whole dataset above into target training set (50\%) and target test test (50\%). 
%The target training dataset is not allowed to overlap with the attack dataset for model inversion attack, which corresponds to the settings behind other black-box MI attacks \cite{yang2019neural}. Then, we adopt the target test dataset for attacking. Typically, we split the target test dataset defined above into attack training set (80\%) and attack test set (20\%). 
% We follow the common setting in black-box model inversion attack\cite{yang2019neural}, where the target training dataset is not allowed to overlap with the attack dataset. To this end, 80\% of the target test dataset is used to train the attack model, while the rest 20\% is used for evaluation.

\textbf{Target models.} 
We implemented different target models for each use case. For the iCV-MEFED emotion task, and identity tasks in iCV-MEFED and CelebA, our target model has 3 convolutional and 3 pooling layers. For the MNIST digit task, the target model has 2 convolutional and 2 pooling layers. Instead of using state-of-the-art deeper models, we limited to smaller deep networks so that CAM explanations are not too small at 16$\times$16 pixels. We use the ADAM optimizer with learning rate $10^{-4}$, $\beta_1=0.5$, $\beta_2=0.999$. 
% blur performance for CNN(Blur): 13.6\% for iCV-MEFED, 0.97\% for CelebA, 19.4\% for MNIST), compared to random guess (0.87\%, 0.1\%, 10\%, respectively).

\textbf{Explanation types.} We evaluated with four popular saliency map XAI types. 
Gradient explanation ($\nabla_{\bm{x}} \bm{y}_t$) \cite{simonyan2013deep} describes the sensitivity of the prediction toward input features. 
By multiplying gradients and input features element-wise, $\text{Gradient} \odot \text{Input}$ ($\nabla_x \bm{y}_t \odot \bm{x}$) \cite{shrikumar2016not} describes the influence of each input feature on the prediction. 
The aforementioned explanations are very fine-grained.
In contrast, Grad-CAM ($\text{ReLU}{\left(\sum_{k}{\alpha_k^c A^k}\right)}$) \cite{selvaraju2017grad} aggregates a weighted sum of activation maps in the final convolutional layer to provide smoother attribution-based saliency maps that are more related to features learned by the CNN. 
% \textcolor{green}{LRP~\cite{bach2015pixel} operates by propagating the prediction backward in the neural network by means of purposely designed local propagation rules.}
The last explanation type we evaluated was layer-wise relevance propagation (LRP)~\cite{bach2015pixel}, which attributes the importance of pixels by backpropagating the relevance of neurons in a neural network while obeying the axiom of conservation of total relevance.
%is another form of explanation, which is composed as a weighted sum of activation maps from the last convolutional layer in the CNN, i.e., the saliency map explanation is $E^c = ReLU{\left(\sum_{k}{\alpha_k^c A^k}\right)}$, where $A^k$ is the $k$th activation map ($k \in K$), and $\alpha_k^c=\frac{1}{HW}\sum_{i,j}^{H,W} \frac{\partial y^c}{\partial A_{ij}^k}$ is the gradient-based importance weight.
Furthermore, we evaluated multiple explanations as Alternative CAMs ($\Sigma$-CAM) and Constituent CAMs ($\partial$-CAM), and XAI-aware attention transfer with surrogate CAM ($\check{\bm{E}_t} = \text{s-CAM}$) and reconstructed surrogate CAM ($\hat{\bm{E}_r} = \text{rs-CAM}$).

\textbf{Baseline inversion attack models.} 
Since there are no prior methods exploiting explanations for model inversion attack, we compared our XAI-aware attack approaches against 
Fredrikson et al.'s original model inversion \cite{fredrikson2015model} (Fredrikson), 
and Yang et al.'s transposed CNN using only the target prediction \cite{yang2019neural} (Prediction only).
%, and Zhang et al.'s use of blurred images as auxiliary data \cite{zhang2020secret} (CNN(Blur))). We blurred the input image with a Gaussian blur ($\sigma=2$ for iCV, and $\sigma=6$ for CelebA and MNIST) to obtain an auxiliary knowledge for comparison. We chose the blur level such that the blurred images do not leak much private information themselves (attack accuracy =

\subsection{Evaluation Metrics}
We evaluated the privacy risk of model inversion attacks quantitatively with multiple metrics to gather multiple evidences of how well the inversion reconstructs the private target image $x$ (i.e., input instance put into the target model) and how well sensitive information $y_s$ can be recovered. 

\textbf{Pixelwise Similarity ($1-MSE_x$).} Mean squared error (MSE) is commonly used to evaluate regression problems. We scale both target and reconstructed images to a unit square, and normalize pixel values to [0,1] and calculate their MSE. The similarity metric, $s=1-MSE$, is image size invariant and increases with image similarity.

\textbf{Image Similarity (SSIM).} MSE does not linearly represent how humans perceive image similarity, so we also employ the perception-based Structural Similarity Index Measure (SSIM) \cite{wang2004image} to evaluate image quality \cite{zhou2019vision}:
\vspace{-0.1cm}
\begin{equation}
     \text{SSIM}\left(x_a, x_b\right) = \frac{\left(2 \mu_a \mu_b + C_{\mu} \right) \left(2 \sigma_{ab} + C_{\sigma} \right)}{\left(\mu_a^2 + \mu_b^2 + C_{\mu} \right) \left(\sigma_a^2 + \sigma_b^2 + C_{\sigma} \right)} ,
\label{eq:SSIM}
\vspace{-0.1cm}
\end{equation}
where $x_a$ and $x_b$ are two images being compared, $\mu_*$ and $\sigma_*$ represents the pixel value mean and standard deviation, respectively,  $C_{\mu} =(K_{\mu} L)^2$ and $C_{\sigma} =(K_{\sigma} L)^2$ are constants to control instability, $L$ is the dynamic range of the pixel values (255 for 8-bit images), and $K_{\mu}=0.01$ and $K_{\sigma}=0.03$ are chosen to be small. To compare images at different levels of granularity, we compare Gaussian kernels for both images at specified standard deviations, $\sigma$ (smaller $\sigma$ for more precise comparison), and compute their mean. We calibrated $\sigma$ to match the human perceived similarity as judged by two co-authors: $\sigma=1.5$ for iCV-MEFED, $\sigma=2.5$ for CelebA, and $\sigma=1.5$ for MNIST.

\textbf{Attack Accuracy.} We trained an Attack Evaluation CNN Model, $M_s$, on the attack task (e.g., face identity) on original images of the full dataset to represent a universal capability to predict sensitive information from original images. If the classifier can correctly label a reconstructed image, then that image leaks private information. The evaluation model is trained on identity (ID) labels for the iCV-MEFED and CelebA datasets, and digit labels for MNIST. Models are not trained on reconstructed images. Model architectures as CNNs are described in supplementary materials. The model accuracy on reconstructed images indicates the loss of privacy due to the model inversion attack; this represents the risk of de-anonymization and identity theft.

\textbf{Attack Embedding Similarity ($e^{-MSE_s}$).} While the aforementioned similarity metrics are data-centric (for images), they are agnostic to the attack task. Poorly reconstructed data can still leak much sensitive information, e.g., a reconstructed face with obfuscated nose and chin can still be recognized well if the eyes or mouth features are preserved. After training the Attack Evaluation CNN Model $M_s$, we compute the feature embedding $\bm{z}$ of the input image $\bm{x}$ from the penultimate layer, and calculate the MSE between the embeddings of the reconstructed and original images, $\bm{z}_r$ and $\bm{z}$, respectively. 
The Attack Embedding Similarity is computed as $e^{-|\bm{z}-\bm{z}_r|_2^2}$; this decreases with distance and is bounded between 0 and 1. Zhang et al. \cite{zhang2020secret} computed Feature Distance 
% \textcolor{green}{as well.} \st{
and KNN Distance metrics between the reconstructed image and centroid of images for the same class; these are quantitative proxies for the classifying the reconstructed image into identity classes, which is what their Attack Accuracy metric measures. Instead, our metric determines how closely the reconstructed image matches the original image in terms of the attack task.
%}%, since the feature embedding space was trained to represent that task.

\subsection{Experiment Results}

We first conducted an ablation study on the face-emotion use case with iCV-MEFED to determine the most performant XAI Input Method, and analyzed the interaction effect between XAI Input Method and XAI Type. We then conducted comparison studies with baselines \cite{fredrikson2015model, yang2019neural}, the best single-explanation XAI-aware inversion model (Flatten+U-Net) with the popular Grad-CAM \cite{selvaraju2017grad} explanation, and its reconstructed, surrogate variant (rs-CAM) for non-explainable target models. For generalization, the latter studies were conducted across multiple datasets.

\subsubsection{Attacking with different XAI Input Methods}

\begin{figure}[t]
    \centering
    \includegraphics[width=8.2cm]{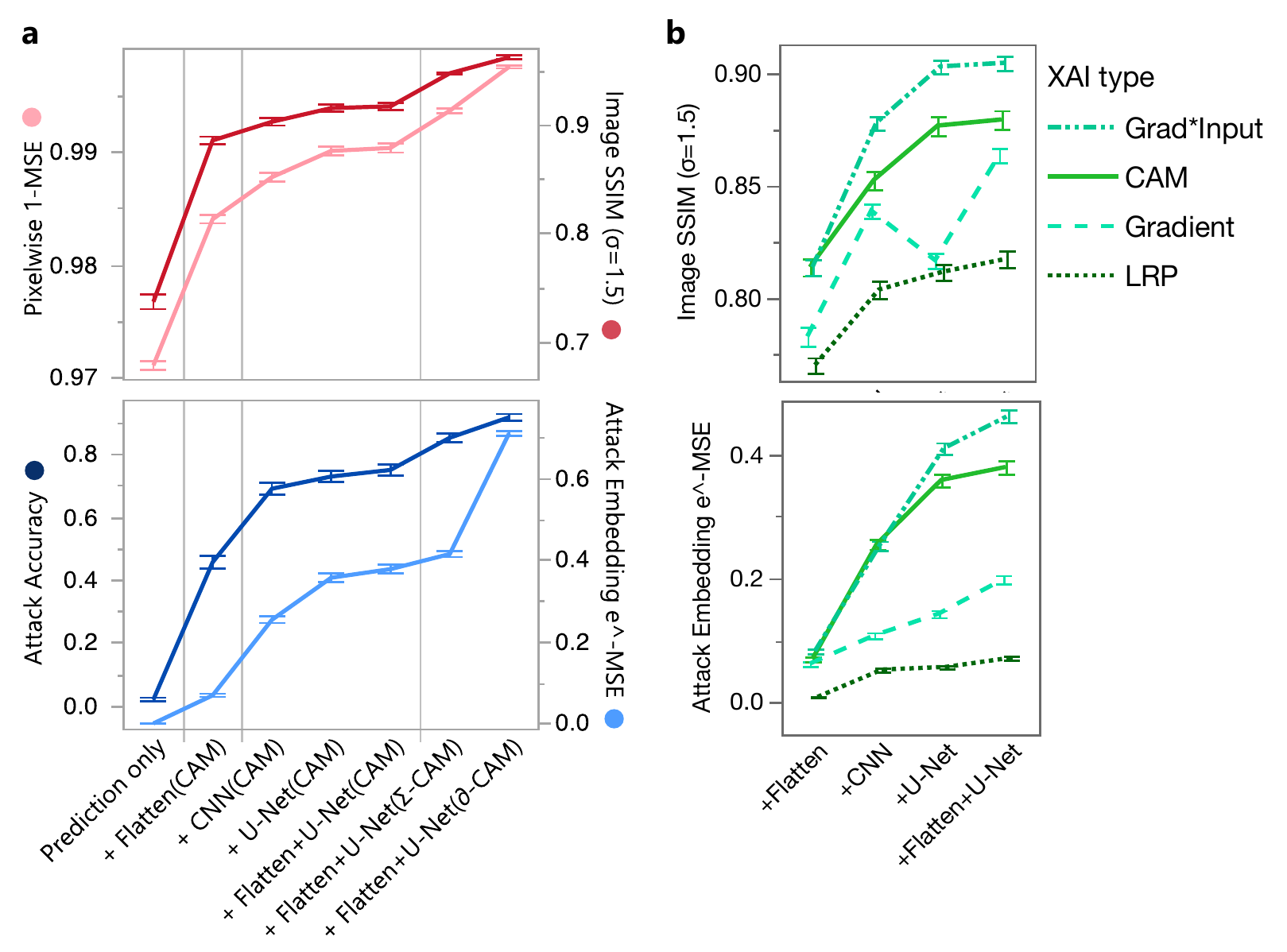}
    \caption{
    Inversion attack performance for different XAI input methods and XAI types
    of iCV-MEFED (Emotion target task)
    increases for 
    a) more spatially-aware architectures and with multi-explanations, and 
    b) explanation types that include sharper details of the input image.
    Error bars indicate 90\% confidence interval.
    }
    \label{fig:xaiInputs}
    \vspace{-.5cm}
\end{figure}

We found that adding more spatially-aware architectures improved the inversion attack performance in the order: Prediction only $<$ Flatten $<$ CNN $<$ U-Net $<$ Flatten+U-Net (see Figure \ref{fig:xaiInputs}a).
As expected, with the least information provided, the Prediction only inversion model had lowest performance in terms of reconstruction similarity and attack accuracy. Exploiting explanation through any method significantly improved the inversion performance. 
However, the Flatten input method achieved the lowest performance among XAI-aware inversion models. This could be due to the lack of spatial information and high dimensionality (e.g., $256\times256$ for CelebA). 
The CNN input method improved inversion performance by performing dimensionality reduction to infer features that are used for model inversion. Although the information is reduced and only the latent embedding is provided to the TCNN module, the learned features are clearly more useful for encoding concepts in the original image to help to improve the inversion attack. However, the TCNN still does not have explicit spatial information from the explanations and remains limited in performance.
On the other hand, by adding bypass connectors, the U-Net architecture is able to leverage on pixel information from the raw information at multiple levels of convolution to learn an inversion function. Hence, U-Net is a more successful architecture than CNN and Flatten. Finally, combining Flatten and U-Net is able to allow the inversion model to acquire raw pixel information and semantic features for a more knowlegeable model inversion attack. Therefore, this provided the strongest attack performance.

% TODO: Try to justify with prior literature.

\subsubsection{Attacking multi-explanation target models}
We evaluated whether providing richer CAM explanations increased inversion performance.
As expected, models that provide more explanations are at greater privacy risk in the order: CAM $<$ $\Sigma$-CAM $<$ $\partial$-CAM. 
Alternative explanations ($\Sigma$-CAM) support contrastive explanations with more information, but this poses further privacy risk. Hence, developers should limit access to too many alternative explanations to limit leakage.
$\partial$-CAMs are useful for model debugging but are unlikely to be provided to end-users to avoid information overload. However, if the API does not restrict access to such explanations, an attacker can discover them and perform very accurate inversion attacks. %almost to the level of having access to very informative blur obfuscated input images (see CNN(Blur) baseline).

\vspace{-1pt}
\subsubsection{Attacking target models with different XAI types}

\begin{figure}[t]
    \centering
    \includegraphics[width=8.2cm]{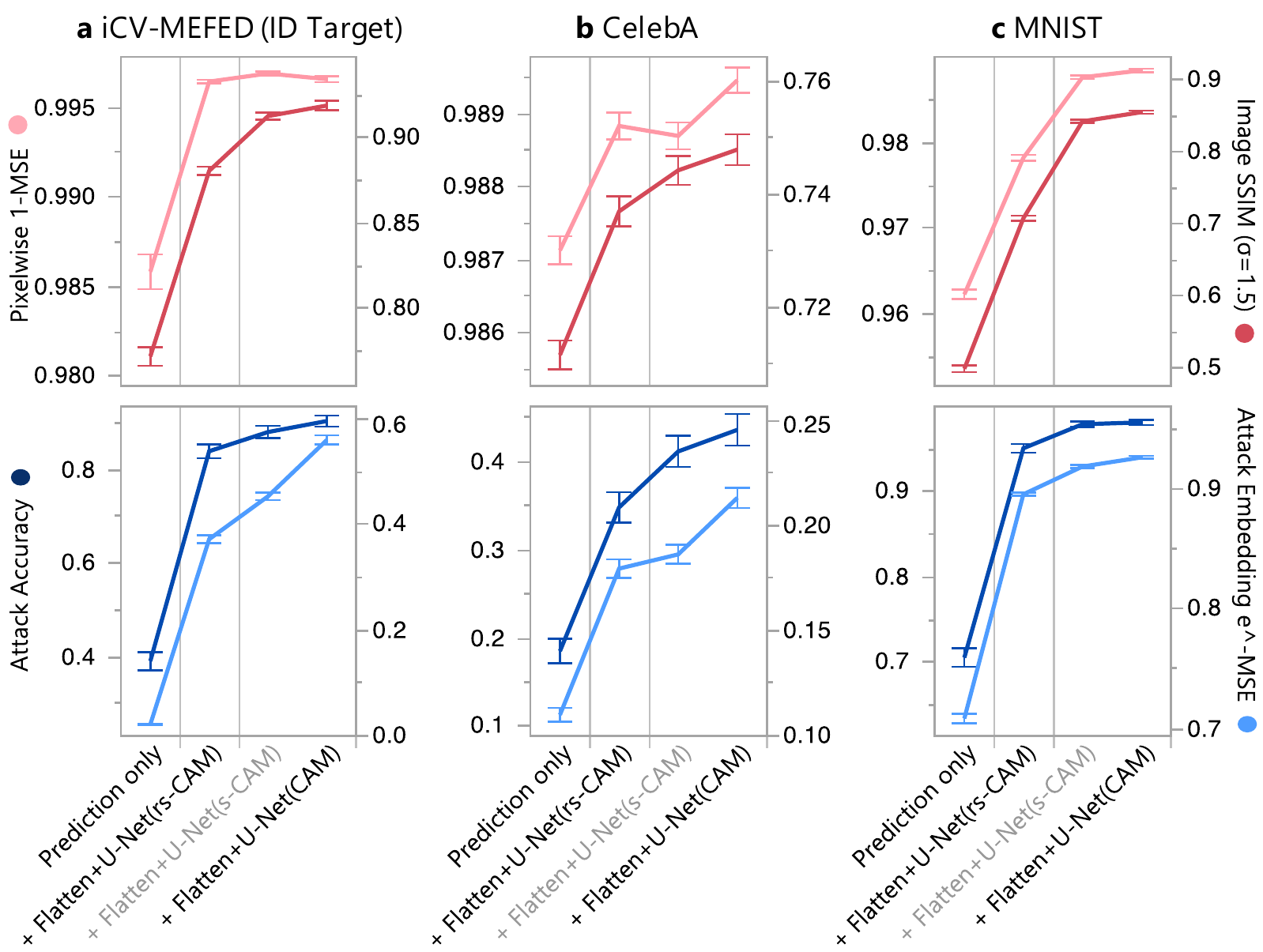}
    \caption{
    Inversion attack performance across different datasets showing increased privacy risk when exploiting target explanations (CAM) and with attention transfer.
    %Performance reported reconstruction similarity (Pixelwise 1-MSE, SSIM, Attack Embedding $e^{-MSE}$) and attack accuracy (e.g., re-identification).
    Even without target explanations, inversion performance with reconstructed, surrogate explanations (rs-CAM) was similar to exploiting target explanations.
    Surrogate explanations (s-CAM) are not available at prediction time and is only shown as intermediate comparison. Error bars indicate 90\% confidence interval.
    %For comparison, XAI-aware inversion attack is almost as performant as inversion with blurred images as auxiliary data.
    Performance with baseline method by Fredrikson et al. \cite{fredrikson2015model} is significantly poorer and reported in supplementary materials.
    }
    \label{fig:xaiAttention}
    \vspace{-.5cm}
\end{figure}

% Reason why Gradient*Input is good: grad is like noise, so discounting noise will give Input, therefore, extremely informative
% Alternative interpretation: grad*input = input with noise, so it is a weakened ground truth for inversion problem

Inversion performance improved in the order: Prediction only $<$ LRP $<$ Gradient $<$ CAM $<$ $\text{Gradient} \odot \text{Input}$. This trend was consistent for different XAI Input Methods (Figure \ref{fig:xaiInputs}b).
LRP and Gradient explanations provide the least information for attack since they only communicate the sensitivity per pixel and do not have direct information about the input image. 
%As LRP utilizes purposely designed local propagation rules to propagate, there is information loss compared with Gradient} \st{LRP and Gradient explanations provide the \st{second} least information for attack since they only communicate the sensitivity per pixel and do not have direct information about the input image.}
In contrast, $\text{Gradient} \odot \text{Input}$ encodes much knowledge about the original image, due to the element-wise multiplication of the Hadamard operator, which the inversion model can more easily learn to recover. 
CAM explanations combine gradient information in its importance weights $\alpha_k^c$ and a transformation of the original image in the activation maps $A^k$ of convolution kernels. Thus, they leak more private information than Gradient, but less than $\text{Gradient} \odot \text{Input}$ because of the weighted average aggregation that obfuscates information. In contrast, Constituent CAMs ($\partial$-CAM) that retain knowledge of individual kernels leak more for inversion attacks (see Figure \ref{fig:xaiInputs}a).

\subsubsection{Attacking non-explainable target models}

%{\color{red}PCC(CAM, s-CAM), PCC(CAM, rs-CAM), PCC(s-CAM, rs-CAM). It was to measure how well the CAM information is retained}

We found that inverting predictions of non-explainable target models with surrogate explanations can increase inversion performance in the order: Prediction only $<$ rs-CAM $<$ s-CAM $<$ CAM (see Figure \ref{fig:xaiAttention}). s-CAM represents the CAM from the surrogate model, and inversion with s-CAM represents an upper bound of the explanation-inversion attack. While rs-CAM performance is slightly lower than CAM, it is significantly higher than Prediction only. 
We further found that attack models trained on out-of-distribution (OOD) data~\cite{ng2014data} can still increase inversion attack performance, albeit weaker (see Supplementary Figure \ref{fig:ood}a).
This demonstrates the significant threat that inversion attack can be made much more aggressive even when target models provide no explanation. This is due to the ability to train more knowledgeable attack models with attention transfer. 
% \textcolor{green}{See Supplementary Figure \ref{fig:ood}a for out-of-distribution (OOD) vs. IID when exploring rs-CAM.} 

% \st{Furthermore, multiple surrogate models can be trained for different prediction tasks, even if the target model only predicts one task; this will provide multiple surrogate explanations. Similar to our results on multiple explanations, we expect that multi-explanation surrogate models can further improve the inversion attack performance. Finally, attention transfer can also be applied to further improve the inversion performance even for explainable target models by complementing with surrogate explanations of other tasks.}

\subsubsection{Impact of explanation quality on attack accuracy}
\begin{figure}[t]
    \centering
    \includegraphics[width=8.2cm]{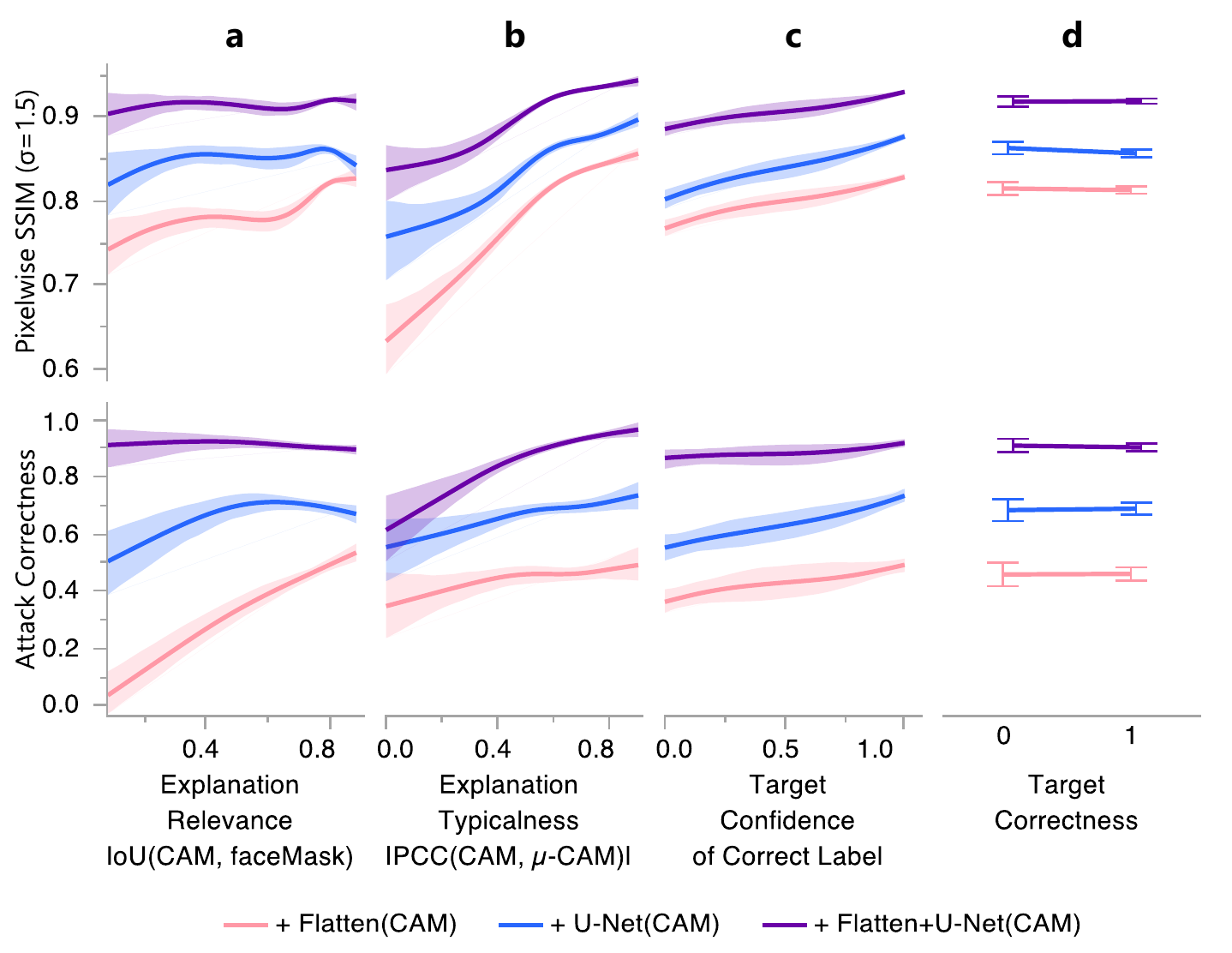}
    \caption{
    Investigation of the influence of target explanation and prediction factors on attack performance.
    Cubic spline fit to data points with 90\% confidence interval.
    }
    \label{fig:relationXaiAttack}
    \vspace{-.3cm}
\end{figure}
We investigated the influence of multiple factors in explanation quality on the inversion attack accuracy (see Figure \ref{fig:relationXaiAttack}), in terms of image reconstruction performance (SSIM) and attack accuracy. This analysis was performed on the iCV-MEFED dataset with the emotion target task, CAM explanation, and re-identification attack task. 

Some CAM explanations highlight irrelevant regions such as the black masked region of the face image. We quantify Explanation Relevance as normalized Intersection over Union (IoU) between the CAM and valid face region (Mask). We found that more relevant explanations improve attack performance, though Flatten+U-Net can achieve high accuracy for less relevant explanations (Figure \ref{fig:relationXaiAttack}a).

Some explanations can be atypical for a class prediction. We quantify Explanation Typicalness as the Pearson correlation coefficient (PCC) between the CAM and the pixelwise average CAM of its class ($\mu$-CAM). We used PCC since it is more appropriate for lower-dimensionality saliency maps \cite{adebayo2018sanity}. We found that attack performance decreased for less typical explanations (Figure \ref{fig:relationXaiAttack}b).

We found a slight effect that the prediction confidence in the target model increased attack performance (Figure \ref{fig:relationXaiAttack}c). Unlike \cite{zhang2020secret} which showed that target model performance is correlated with attack performance, we found that target model accuracy (Figure \ref{fig:relationXaiAttack}d) did not affect attack performance. This apparent contradiction is due to differing attack objectives: \cite{zhang2020secret} sought to invert the prototypical image representing a class label to leak knowledge about the training dataset, while we sought to invert the original image of a test instance. Our objective is similar to \cite{yang2019neural}.

% \section{Discussion and Conclusion}
\section{Conclusion}
We have presented several methods for model inversion attack to exploit model explanations to demonstrate increased privacy risk. 
This highlights the conflict between explainability and privacy, and is a critical first step towards finding an optimal balance between these two requirements for responsible AI.
Our approach trains a multi-modal transposed CNN with a Flatten input layer and U-Net architecture to acquire detailed information of the explanation, lower-dimension semantic concepts from latent features, and multi-scale spatial information from bypass connectors. 
With this as the core XAI-aware inversion model, we further train a meta-architecture to increase the inversion performance even against non-explainable target models. This approach trains an explainable surrogate target model, then trains an explanation inversion model from the target predictions, which reconstructs an explanation to be used as a surrogate input for the XAI-aware inversion model. 
% Attack accuracies
% - iCV Emotion: Prediction only: 0.022135, Flatten+U-Net(CAM): 0.752083 (33.4x), Flatten+U-Net(A-CAM): 0.854583 (38.6x), Flatten+U-Net(C-CAM): 0.919583 (41.5x)
% - iCV ID: Fredrikson: 0.070417, Prediction only: 0.390417, Flatten+U-Net(rs-CAM): 0.839167 (2.15x), Flatten+U-Net(CAM): 0.903750 (2.31x)
% - CelebA ID: Fredrikson: 0.009073, Prediction only: 0.185484, Flatten+U-Net(rs-CAM): 0.348505 (1.88x), Flatten+U-Net(CAM): 0.436492 (2.35x)
% - MNIST: Fredrikson: 0.105361, Prediction only: 0.705295, Flatten+U-Net(rs-CAM): 0.951118 (1.35x), Flatten+U-Net(CAM): 0.981481 (1.39x)
Our experiment results show an increased attack accuracy when exploiting target explanations (up to 33x for iCV-MEFED emotion task, and 2.4 for CelebA ID task), even higher for multi-explanations such as contrastive or detailed constituent explanations (up to 39x, 42x for iCV emotion task, respectively), and, concerningly, even without target explanations (with surrogate, up to 2.15x). We found that activation-based (saliency map) explanations leak more privacy than sensitivity-based (gradients) explanations, resulting in higher inversion performance, and so do explanations that are more relevant and typical.
% \textcolor{green}{Our study not only is critical first step to dissect how different XAI forms leak privacy, but also provides new solutions. E.g., specific privacy defence methods can be developed, through selective incorporation of explanation features.}
For future work, we will the inversion attack approach can be extended for different explanations (e.g., feature visualizations, concept activation vectors), different data modalities (e.g., spectrograms of audio) and investigate techniques for privacy defense.

\section*{Acknowledgement}
\noindent
% \begin{acks}
The research is supported by the National Research Foundation Prime Minister’s Office, Singapore under its Strategic Capability Research Centers Funding Initiative, the N-CRiPT research center at the National University of Singapore. The Titan Xp used for this research was donated by the NVIDIA Corporation.
% \end{acks}

{\small
\bibliographystyle{ieee_fullname}
\bibliography{references}
}

\clearpage
\appendix
\renewcommand{\figurename}{Supplementary Figure}
\renewcommand{\tablename}{Supplementary Table}
% \thispagestyle{empty}

% \section{Supplementary \#1}
\section{Supplementary method}
% Previously, we have described the general approach for model inversion attack models in Figure \ref{fig:models}. In this supplementary section, we will describe general approach for XAI input inversion attack models, detailed approach for inversion attack models, detailed neural network layers for target and inversion attack models.

\subsection{XAI-Aware Inversion Attack Models}
We describe the architectures of our proposed models. 
Supplementary Figure \ref{fig:xaiInputs_high_level} illustrates the meta-architectures for all XAI Input Methods for inversion attacks. 
Supplementary Figure \ref{fig:xaiInputs_detail} illustrates the detailed neural network layer architectures of key inversion models. 
Supplementary Tables \ref{tab:target}-\ref{tab:flatten_unet} describe details of the layer settings for various target and attack models, describing convolutional layers (conv), max pooling layers (pool), fully connected layers (fc), and transposed convolutional layers with large strides (upsample). 
%Each conv layer are followed by a max pooling layer with 2$\times$2 stride.

\vspace{-.3cm}
\setcounter{figure}{0}
\begin{figure}[ht]
    \centering
    \includegraphics[width=8.1cm]{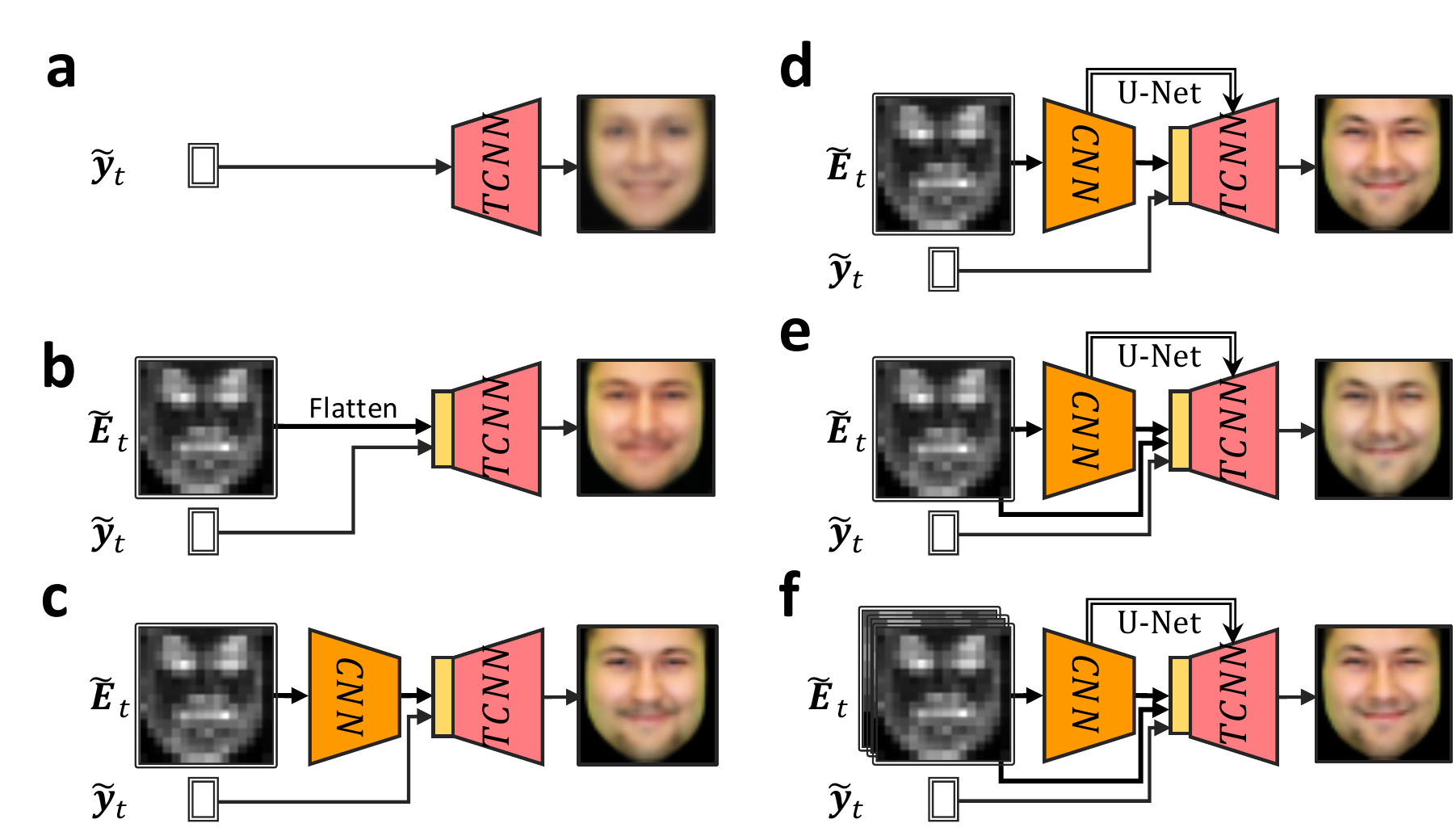}
    \caption{
    Architectures of XAI input inversion attack models:
    % a) Baseline threat model. % Emotion prediction confidences $\tilde{\bm{y}}_t$ are input to a transposed CNN (TCNN) for inversion attack. 
    % b)-f) XAI-aware multi-modal inversion attack models that inputs $\tilde{\bm{E}}_t$ via different input architectures.
    % b) Flattened $\tilde{\bm{E}}_t$ concatenated with $\tilde{\bm{y}}_t$, 
    % c) CNN processing $\tilde{\bm{E}}_t$ concatenated with $\tilde{\bm{y}}_t$, 
    % d) U-Net processing $\tilde{\bm{E}}_t$ concatenated with $\tilde{\bm{y}}_t$,  
    % e) Combined Flatten and U-Net processing $\tilde{\bm{E}}_t$ concatenated with $\tilde{\bm{y}}_t$,
    % f) Combined Flatten and U-Net processing multiple explanations.
    % % XAI-aware multi-modal inversion attack model that inputs $\tilde{\bm{E}}_t$ via architecture combining Flatten and U-Net when exploiting multiple f) explanations.
    a) Baseline threat model,
    b) Flattened $\tilde{\bm{E}}_t$ concatenated with $\tilde{\bm{y}}_t$,
    c) CNN for dimensionality reduction,
    d) U-Net for dimensionality reduction and spatial knowledge,
    e) Combined Flatten+U-Net,
    f) Combined Flatten+U-Net on multiple explanations as a 3D tensor.
    }
    \label{fig:xaiInputs_high_level}
    \vspace{-.3cm}
\end{figure}

\newpage
\vspace{-.3cm}
\begin{figure}[h]
    \centering
    \hspace{-.4cm}
    \includegraphics[width=8.4cm]{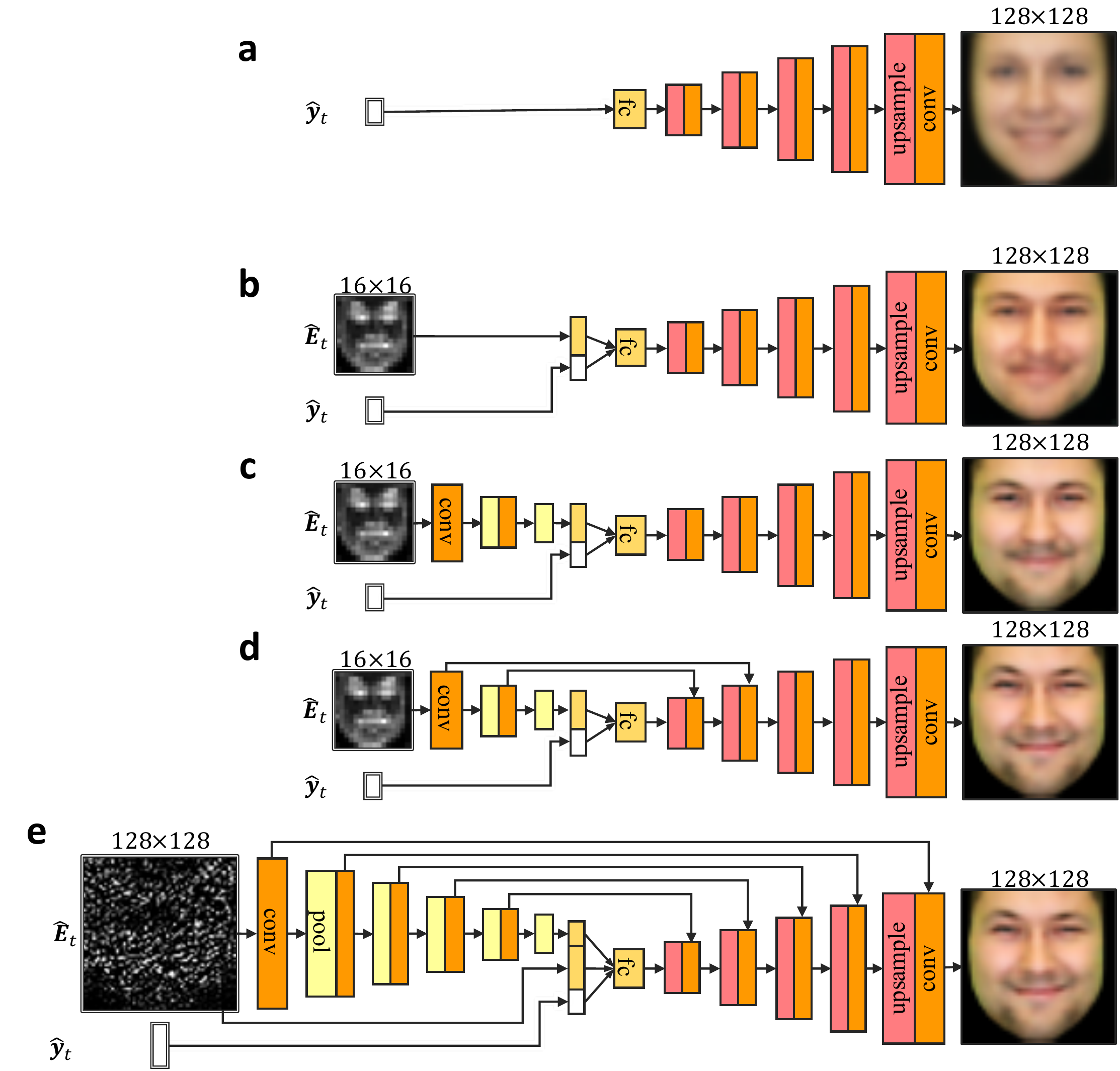}
    \caption{
    % Detailed architectures of XAI input inversion attack models. 
    % a) Baseline threat model. Emotion prediction confidences $\tilde{\bm{y}}_t$ are input to multiple upsample layers followed by conv layers for inversion attack. 
    % XAI-aware multi-modal inversion attack model that inputs CAM $\tilde{\bm{E}}_t$ via different input architectures b)-d):
    % b) Flattened $\tilde{\bm{E}}_t$ concatenated with $\tilde{\bm{y}}_t$, 
    % c) CNN processing $\tilde{\bm{E}}_t$ by conv and pool layers and then concatenated with $\tilde{\bm{y}}_t$, 
    % d) U-Net processing $\tilde{\bm{E}}_t$ by conv and pool layers, then concatenated with $\tilde{\bm{y}}_t$. Meanwhile, conv layers processing CAM are bypass connected with conv following upsample layers. There is no bypass connections for few conv at last because CAM size 8x8 is smaller than image size 64x64.
    % XAI-aware multi-modal inversion attack model that inputs Gradient $\tilde{\bm{E}}_t$ via architecture combining Flatten and U-Net in e). As Gradient size is 64x64, $\tilde{\bm{E}}_t$ is processed by more conv and pool layers, and then more bypass connections are introduced.
    Detailed architectures of inversion attack models for different XAI Input Methods. 
    % a) Baseline threat model. Emotion prediction confidences $\tilde{\bm{y}}_t$ are input to multiple upsample-convolution block.
    a) Prediction only TCNN with alternating transposed conv (upsample) and conv layers from $\tilde{\bm{y}}_t$ only.
    % XAI-aware multi-modal inversion attack model that inputs CAM $\tilde{\bm{E}}_t$ via different input architectures (b-d):
    % b) Flattened $\tilde{\bm{E}}_t$ concatenated with $\tilde{\bm{y}}_t$, 
    b) Flatten(CAM) method first flattens the explanation $\tilde{\bm{E}}_t$ pixels as a 1D vector and concatenates with $\tilde{\bm{y}}_t$, then upsamples with the same TCNN layers as (a).
    % c) $\tilde{\bm{E}}_t$ processed by conv and pool layers in CNN and then concatenated with $\tilde{\bm{y}}_t$, 
    c) CNN(CAM) method reduces the dimension in $\tilde{\bm{E}}_t$ with several conv and pool layers to produce a 1D embedding vector that is concatenated with $\tilde{\bm{y}}_t$.
    % d) $\tilde{\bm{E}}_t$ processed by conv and pool layers in U-Net, then concatenated with $\tilde{\bm{y}}_t$. Meanwhile, bypass connections between conv and upsample layers help to maintain spatial knowledge. There is no bypass connection for last conv layers due to the dimensionality difference between CAMs and images.
    d) U-Net(CAM) method adds to CNN method with bypass connectors from each conv layer in the CNN to corresponding conv layer in the TCNN. For CAM explanations which are smaller than images 
    % (\textcolor{green}{16$\times$16 vs 128$\times$128})
    , there are no bypass connectors for later larger TCNN layers.
    % e) Gradient $\tilde{\bm{E}}_t$ processed by combined Flatten and U-Net. As the Gradient has the same size with the image, balanced bypass connections are introduced between all conv and upsample layers.
    e) Flatten+U-Net(Gradient) method with more conv and pool layers due to the larger pixel size of the gradient explanation, with 1D embedding concatenated with Flatten(Gradient) and $\tilde{\bm{y}}_t$. Since the gradient explanation $\tilde{\bm{E}}_t$ and reconstructed image output are in the same size 
    % (\textcolor{green}{128$\times$128})
    , the U-Net has all bypass connectors.
    }
    \label{fig:xaiInputs_detail}
    % \vspace{-.3cm}
\end{figure}

% \newpage
\clearpage
% \pagebreak

% \subsection{The detailed neural network layers for target and XAI inversion attack models}
% We describe the detailed information of neural network layers for target models in Supplementary Table \ref{tab:target_mnist}, \ref{tab:target}, and inversion attack models in Supplementary Table \ref{tab:flatten}, \ref{tab:cnn}, \ref{tab:unet}, \ref{tab:grad}. \footnote{As the stride of pool layers is 2x2 across all models, we do not present it in tables.}

% Unlike \cite{zhang2020secret} that trained deeper CNN models (e.g., VGG, ResNet), we trained shallower CNNs for the target model since the test images are small (64x64) and we want CAM explanations that are informative and not too small (8x8). The target model introduces 2 blocks for handwriting digit recognition in MNIST \cite{lecun1998gradient} in Supplementary Table \ref{tab:target_mnist} and 3 blocks for emotion and identity recognition in iCV-MEFED \cite{loob2017dominant}, identity recognition in CelebA \cite{liu2018large} in Supplementary Table \ref{tab:target}. Corresponding to various inversion attack models in Supplementary Figure \ref{fig:xaiInputs_detail}, we provide detailed neural network layers from Supplementary Table \ref{tab:flatten} to Supplementary Table \ref{tab:grad}. As pool layer is normally set with stride 2x2, we do not present in these tables.

% \vspace{-0.5cm}
\begin{table}[ht]
% \small
\footnotesize
\centering
\begin{tabular}{cccccc}
\hline
Type  & Kernel & Stride & Padding & \multicolumn{1}{c}{\begin{tabular}[c]{@{}c@{}}Feature\\ Map\end{tabular}} & Outputs \\ \hline
input &     &     &     & 128$\times$128   & 1      \\ \hline
conv & 3$\times$3    & 1$\times$1    & 1    & 128$\times$128   & 128      \\ \hline
pool & 2$\times$2     & 2$\times$2    & 0    &  64$\times$64   & 128      \\ \hline
conv & 3$\times$3    & 1$\times$1    & 1    & 64$\times$64   & 256     \\ \hline
pool & 2$\times$2     & 2$\times$2    & 0     &  32$\times$32  & 256        \\ \hline
conv & 3$\times$3    & 1$\times$1    & 1    & 32$\times$32     & 512    \\ \hline
pool & 2$\times$2     & 2$\times$2    & 0     & 16$\times$16     & 512     \\ \hline
fc   &        &        &       &  & 512    \\ \hline
fc    &        &        &       &   & $|C|$    \\ \hline
\end{tabular}
% \captionsetup{font={small}}
\vspace{0.3cm}
\caption{Network layers for iCV-MEFED target models. $|C|$ is the number of classes.}
\label{tab:target}
\end{table}

% \vspace{-0.5cm}
\begin{table}[ht]
% \small
\footnotesize
\centering
\begin{tabular}{cccccc}
\hline
Type  & Kernel & Stride & Padding & \multicolumn{1}{c}{\begin{tabular}[c]{@{}c@{}}Feature\\ Map\end{tabular}} & Outputs \\ \hline
input &     &     &     & 256$\times$256   & 1      \\ \hline
conv & 3$\times$3    & 1$\times$1    & 1    & 256$\times$256   & 128      \\ \hline
pool & 2$\times$2     & 2$\times$2    & 0    &  128$\times$128   & 128      \\ \hline
conv & 3$\times$3    & 1$\times$1    & 1    & 128$\times$128   & 256     \\ \hline
pool & 2$\times$2     & 2$\times$2    & 0     &  64$\times$64  & 256        \\ \hline
conv & 3$\times$3    & 1$\times$1    & 1    & 64$\times$64     & 512    \\ \hline
pool & 2$\times$2     & 2$\times$2    & 0     & 32$\times$32     & 512     \\ \hline
conv & 3$\times$3    & 1$\times$1    & 1    & 32$\times$32     & 1024    \\ \hline
pool & 2$\times$2     & 2$\times$2    & 0     & 16$\times$16     & 1024     \\ \hline
fc   &        &        &       &  & 1024    \\ \hline
fc    &        &        &       &   & $|C|$    \\ \hline
\end{tabular}
% \captionsetup{font={small}}
\vspace{0.3cm}
\caption{Network layers for CelebA target models. $|C|$ is the number of classes.}
\label{tab:target}
\end{table}

% \vspace{-0.3cm}
\begin{table}[h]
% \small
\footnotesize
\centering
\begin{tabular}{cccccc}
\hline
Type  & Kernel & Stride & Padding & \multicolumn{1}{c}{\begin{tabular}[c]{@{}c@{}}Feature\\ Map\end{tabular}} & Outputs\\ \hline
input &     &     &     & 32$\times$32   & 1      \\ \hline
conv & 3$\times$3    & 1$\times$1    & 1    & 32$\times$32   & 128    \\ \hline
pool & 2$\times$2     & 2$\times$2    & 0    & 16$\times$16   & 128     \\ \hline
conv & 3$\times$3    & 1$\times$1    & 1     & 16$\times$16  & 256     \\ \hline
pool & 2$\times$2     & 2$\times$2   & 0    & 8$\times$8   & 256    \\ \hline
fc   &        &        &       &  & 512    \\ \hline
fc    &        &        &       &   & $|C|$    \\ \hline
\end{tabular}
% \captionsetup{font={small}}
\vspace{0.3cm}
\caption{Network layers for MNIST target model.}
\label{tab:target_mnist}
\end{table}

% \vspace{-0.3cm}
\newpage
\vspace{-15cm}
\begin{table}[h]
% \small
\footnotesize
\centering
\begin{tabular}{cccccc}
\hline
Type       & Kernel & Stride & Padding & \multicolumn{1}{c}{\begin{tabular}[c]{@{}c@{}}Feature\\ Map\end{tabular}} & Outputs\\ \hline
input &     &     &     &    & $|C|$      \\ \hline
fc     &        &        &      &    & $|C|$     \\ \hline
upsample & 4$\times$4    & 1$\times$1    & 0   & 4$\times$4    & 1024    \\ \hline
conv      & 3$\times$3    & 1$\times$1    & 1   & 4$\times$4    & 1024     \\ \hline
upsample & 4$\times$4    & 2$\times$2    & 1    & 8$\times$8    & 512     \\ \hline
conv      & 3$\times$3    & 1$\times$1    & 1   & 8$\times$8    & 512      \\ \hline
upsample & 4$\times$4    & 2$\times$2    & 1    & 16$\times$16   & 256      \\ \hline
conv      & 3$\times$3    & 1$\times$1    & 1    & 16$\times$16    & 256     \\ \hline
upsample & 4$\times$4    & 2$\times$2    & 1    & 32$\times$32    & 128     \\ \hline
conv      & 3$\times$3    & 1$\times$1    & 1   & 32$\times$32     & 128     \\ \hline
upsample & 4$\times$4    & 2$\times$2    & 1    & 64$\times$64    & 64     \\ \hline
conv      & 3$\times$3    & 1$\times$1    & 1   & 64$\times$64     & 64     \\ \hline
upsample & 4$\times$4    & 2$\times$2    & 1    & 128$\times$128    & 1      \\ \hline
conv      & 3$\times$3    & 1$\times$1    & 1   & 128$\times$128     & 1      \\ \hline
\end{tabular}
% \captionsetup{font={small}}
\vspace{0.3cm}
\caption{Network layers for Prediction only inversion attack model on iCV-MEFED(Supplementary Figure \ref{fig:xaiInputs_detail}a).}
\label{tab:pred_only}
\end{table}

% \newpage
% \vspace{-0.5cm}
% \vspace{-15cm}
\begin{table}[h]
% \small
\footnotesize
\centering
\begin{tabular}{cccccc}
\hline
Type       & Kernel & Stride & Padding & \multicolumn{1}{c}{\begin{tabular}[c]{@{}c@{}}Feature\\ Map\end{tabular}} & Outputs \\ \hline
% input pred    &        &        &       & 1$\times$1  & 6      \\ \hline
% input xAI     &        &        &       & 8$\times$8  & 1     \\ \hline
input($\tilde{\bm{y}_t}$) &     &     &     &    & $|C|$      \\ \hline
input($\tilde{\bm{E}_t}$) &     &     &     & 16$\times$16   & 1      \\ \hline
fc($\tilde{\bm{y}_t}$,$\tilde{\bm{E}_t}$)     &        &        &       & 1$\times$1  & $|C|+16^2$      \\ \hline
upsample & 4$\times$4    & 1$\times$1    & 0   & 4$\times$4    & 1024    \\ \hline
conv      & 3$\times$3    & 1$\times$1    & 1   & 4$\times$4    & 1024     \\ \hline
upsample & 4$\times$4    & 2$\times$2    & 1    & 8$\times$8    & 512     \\ \hline
conv      & 3$\times$3    & 1$\times$1    & 1   & 8$\times$8    & 512      \\ \hline
upsample & 4$\times$4    & 2$\times$2    & 1    & 16$\times$16   & 256      \\ \hline
conv      & 3$\times$3    & 1$\times$1    & 1    & 16$\times$16    & 256     \\ \hline
upsample & 4$\times$4    & 2$\times$2    & 1    & 32$\times$32    & 128     \\ \hline
conv      & 3$\times$3    & 1$\times$1    & 1   & 32$\times$32     & 128     \\ \hline
upsample & 4$\times$4    & 2$\times$2    & 1    & 64$\times$64    & 64     \\ \hline
conv      & 3$\times$3    & 1$\times$1    & 1   & 64$\times$64     & 64     \\ \hline
upsample & 4$\times$4    & 2$\times$2    & 1    & 128$\times$128    & 1      \\ \hline
conv      & 3$\times$3    & 1$\times$1    & 1   & 128$\times$128     & 1      \\ \hline
\end{tabular}
% \vspace{0.3cm}
% \captionsetup{font={small}}
\vspace{0.3cm}
\caption{Network layers for Flatten(CAM) inversion attack model on iCV-MEFED(Supplementary Figure \ref{fig:xaiInputs_detail}b).}
\label{tab:flatten}
\end{table}

% \vspace{-15cm}
\begin{table}[h]
% \small
\footnotesize
\centering
\begin{tabular}{cccccc}
\hline
Type       & Kernel & Stride & Padding & \multicolumn{1}{c}{\begin{tabular}[c]{@{}c@{}}Feature\\ Map\end{tabular}} & Outputs \\ \hline
input($\tilde{\bm{y}_t}$) &     &     &     &    & $|C|$      \\ \hline
input($\tilde{\bm{E}_t}$) &     &     &     & 16$\times$16   & 1      \\ \hline
conv($\tilde{\bm{E}_t}$) & 3$\times$3    & 1$\times$1     & 1   & 16$\times$16     & 256     \\ \hline
pool & 2$\times$2     & 2$\times$2    & 0    & 8$\times$8    & 256      \\ \hline
conv & 3$\times$3    & 1$\times$1    & 1   & 8$\times$8    & 512     \\ \hline
pool & 2$\times$2     & 2$\times$2    & 0    & 4$\times$4    & 512      \\ \hline
conv & 3$\times$3    & 1$\times$1    & 1   & 4$\times$4    & 1024     \\ \hline
pool & 2$\times$2     & 2$\times$2    & 0    & 2$\times$2    & 1024      \\ \hline
fc     &        &        &         &       & 64    \\ \hline
fc($\tilde{\bm{y}_t}$,conv)     &        &        &         &       & $|C|+64$   \\ \hline
upsample & 4$\times$4    & 1$\times$1    & 0  & 4$\times$4    & 1024    \\ \hline
conv      & 3$\times$3    & 1$\times$1    & 1   & 4$\times$4    & 1024     \\ \hline
upsample & 4$\times$4    & 2$\times$2    & 1    & 8$\times$8    & 512     \\ \hline
conv      & 3$\times$3    & 1$\times$1    & 1   & 8$\times$8    & 512      \\ \hline
upsample & 4$\times$4    & 2$\times$2    & 1    & 16$\times$16   & 256      \\ \hline
conv      & 3$\times$3    & 1$\times$1    & 1    & 16$\times$16    & 256     \\ \hline
upsample & 4$\times$4    & 2$\times$2    & 1    & 32$\times$32    & 128     \\ \hline
conv      & 3$\times$3    & 1$\times$1    & 1   & 32$\times$32     & 128     \\ \hline
upsample & 4$\times$4    & 2$\times$2    & 1    & 64$\times$64    & 64      \\ \hline
conv      & 3$\times$3    & 1$\times$1    & 1   & 64$\times$64     & 64      \\ \hline
upsample & 4$\times$4    & 2$\times$2    & 1    & 128$\times$128    & 1      \\ \hline
conv      & 3$\times$3    & 1$\times$1    & 1   & 128$\times$128     & 1      \\ \hline
\end{tabular}
\vspace{0.3cm}
\caption{Network layers for CNN(CAM) inversion attack model on iCV-MEFED(Supplementary Figure \ref{fig:xaiInputs_detail}c).}
\label{tab:cnn}
\end{table}

% \vspace{-5.5cm}
% \begin{table*}[t]
\begin{table}[t]
% \small
\footnotesize
\centering
\begin{tabular}{cccccc}
\hline
Type       & Kernel & Stride & Padding & \multicolumn{1}{c}{\begin{tabular}[c]{@{}c@{}}Feature\\ Map\end{tabular}} & Outputs \\ \hline
input($\tilde{\bm{y}_t}$) &     &     &     &    & $|C|$      \\ \hline
input($\tilde{\bm{E}_t}$) &     &     &     & 16$\times$16   & 1      \\ \hline
conv($\tilde{\bm{E}_t}$) & 3$\times$3    & 1$\times$1     & 1   & 16$\times$16     & 256     \\ \hline
pool & 2$\times$2     & 2$\times$2    & 0    & 8$\times$8    & 256      \\ \hline
conv & 3$\times$3    & 1$\times$1    & 1   & 8$\times$8    & 512     \\ \hline
pool & 2$\times$2     & 2$\times$2    & 0    & 4$\times$4    & 512      \\ \hline
conv & 3$\times$3    & 1$\times$1    & 1   & 4$\times$4    & 1024     \\ \hline
pool & 2$\times$2     & 2$\times$2    & 0    & 2$\times$2    & 1024      \\ \hline
fc     &        &        &         &       & 64   \\ \hline
fc($\tilde{\bm{y}_t}$,conv)     &        &        &         &       & $|C|+64$   \\ \hline
upsample & 4$\times$4    & 1$\times$1    & 0  & 4$\times$4    & 1024    \\ \hline
conv*      & 3$\times$3    & 1$\times$1    & 1   & 4$\times$4    & 1024     \\ \hline
upsample & 4$\times$4    & 2$\times$2    & 1    & 8$\times$8    & 512     \\ \hline
conv*      & 3$\times$3    & 1$\times$1    & 1   & 8$\times$8    & 512      \\ \hline
upsample & 4$\times$4    & 2$\times$2    & 1    & 16$\times$16   & 256      \\ \hline
conv*      & 3$\times$3    & 1$\times$1    & 1    & 16$\times$16    & 256     \\ \hline
upsample & 4$\times$4    & 2$\times$2    & 1    & 32$\times$32    & 128     \\ \hline
conv      & 3$\times$3    & 1$\times$1    & 1   & 32$\times$32     & 128     \\ \hline
upsample & 4$\times$4    & 2$\times$2    & 1    & 64$\times$64    & 64     \\ \hline
conv      & 3$\times$3    & 1$\times$1    & 1   & 64$\times$64     & 64     \\ \hline
upsample & 4$\times$4    & 2$\times$2    & 1    & 128$\times$128    & 1      \\ \hline
conv      & 3$\times$3    & 1$\times$1    & 1   & 128$\times$128     & 1      \\ \hline
\end{tabular}
\vspace{0.3cm}
% \vspace{0.1cm}
\caption{Network layers for U-Net(CAM) inversion attack model. 
conv* indicates connected via bypass connector from CNN conv of same feature map size.
Refer to Supplementary Figure \ref{fig:xaiInputs_detail}d for details of bypass connectors. 
}
\label{tab:unet}
\end{table}
% \end{table*}
% \pagebreak
% \clearpage
% \vspace{-0.5cm}
\vspace{-15cm}
% \begin{table*}[t]
\begin{table}[t]
% \small
\footnotesize
\centering
\begin{tabular}{cccccc}
\hline
Type       & Kernel & Stride & Padding & \multicolumn{1}{c}{\begin{tabular}[c]{@{}c@{}}Feature\\ Map\end{tabular}} & Outputs \\ \hline
input($\tilde{\bm{y}_t}$) &     &     &     &    & $|C|$      \\ \hline
input($\tilde{\bm{E}_t}$) &     &     &     & 128$\times$128   & 1      \\ \hline
conv($\tilde{\bm{E}_t}$) & 3$\times$3    & 1$\times$1    & 1      & 128$\times$128  & 1     \\ \hline
pool & 2$\times$2     & 2$\times$2    & 0    & 64$\times$64  & 1     \\ \hline
conv& 3$\times$3    & 1$\times$1    & 1      & 64$\times$64  & 64     \\ \hline
pool & 2$\times$2     & 2$\times$2    & 0    & 32$\times$32  & 64     \\ \hline
conv& 3$\times$3    & 1$\times$1    & 1      & 32$\times$32  & 128     \\ \hline
pool & 2$\times$2     & 2$\times$2    & 0    & 16$\times$16  & 128     \\ \hline
conv & 3$\times$3    & 1$\times$1    & 1     & 16$\times$16  & 256     \\ \hline
pool & 2$\times$2     & 2$\times$2    & 0    & 8$\times$8  & 256     \\ \hline
conv & 3$\times$3    & 1$\times$1    & 1     & 8$\times$8  & 512     \\ \hline
pool & 2$\times$2     & 2$\times$2    & 0    & 4$\times$4  & 512     \\ \hline
conv & 3$\times$3    & 1$\times$1    & 1     & 4$\times$4  & 1024     \\ \hline
pool & 2$\times$2     & 2$\times$2    & 0    & 2$\times$2    & 1024      \\ \hline
fc     &        &        &       &  & 64      \\ \hline
fc($\tilde{\bm{y}_t}$,conv,     &        &        &       &  & $|C|+64+$     \\ 
  $\tilde{\bm{E}_t}$)  &        &        &       &  & $128^2$     \\ \hline
upsample & 4$\times$4    & 1$\times$1    & 0      & 4$\times$4    & 1024    \\ \hline
conv*      & 3$\times$3    & 1$\times$1    & 1   & 4$\times$4    & 1024     \\ \hline
upsample & 4$\times$4    & 2$\times$2    & 1    & 8$\times$8    & 512     \\ \hline
conv*      & 3$\times$3    & 1$\times$1    & 1   & 8$\times$8    & 512      \\ \hline
upsample & 4$\times$4    & 2$\times$2    & 1    & 16$\times$16   & 256      \\ \hline
conv*      & 3$\times$3    & 1$\times$1    & 1    & 16$\times$16    & 256     \\ \hline
upsample & 4$\times$4    & 2$\times$2    & 1    & 32$\times$32    & 128     \\ \hline
conv*      & 3$\times$3    & 1$\times$1    & 1   & 32$\times$32     & 128     \\ \hline
upsample & 4$\times$4    & 2$\times$2    & 1    & 64$\times$64    & 64     \\ \hline
conv*      & 3$\times$3    & 1$\times$1    & 1   & 64$\times$64     & 64     \\ \hline
upsample & 4$\times$4    & 2$\times$2    & 1    & 128$\times$128    & 1      \\ \hline
conv*      & 3$\times$3    & 1$\times$1    & 1   & 128$\times$128     & 1      \\ \hline
\end{tabular}
\vspace{0.3cm}
% \vspace{0.1cm}
\caption{Network layers Flatten+U-Net(Gradient) inversion attack model.
conv* indicates connected via bypass connector from CNN conv of same feature map size.
Refer to Supplementary Figure \ref{fig:xaiInputs_detail}e for details of bypass connectors. 
}
\label{tab:flatten_unet}
% \end{table*}
\end{table}

% \pagebreak
\clearpage
% \section{Supplementary \#2}
\section{Supplementary results}

This section adds results with other metrics, from the Fredrikson inversion model \cite{fredrikson2015model} baseline and more demonstration instances for each dataset.

\subsection{Attacking with different XAI Input Methods}

% Supplementary Figure \ref{fig:emo_target} demonstrates explanations and image reconstructions for all XAI Input Methods for more demonstration instances.

% As supplement to Figure 4b in the paper, we include results for Pixelwise Similarity $1-\text{MSE}_x$ and Attack Embedding Similarity $e^{-\text{MSE}_x}$, along with Image Similarity SSIM and Attack Accuracy, in Supplementary Figure \ref{fig:xairelation}.

\begin{figure}[h]
    \centering
    \includegraphics[width=8.4cm]{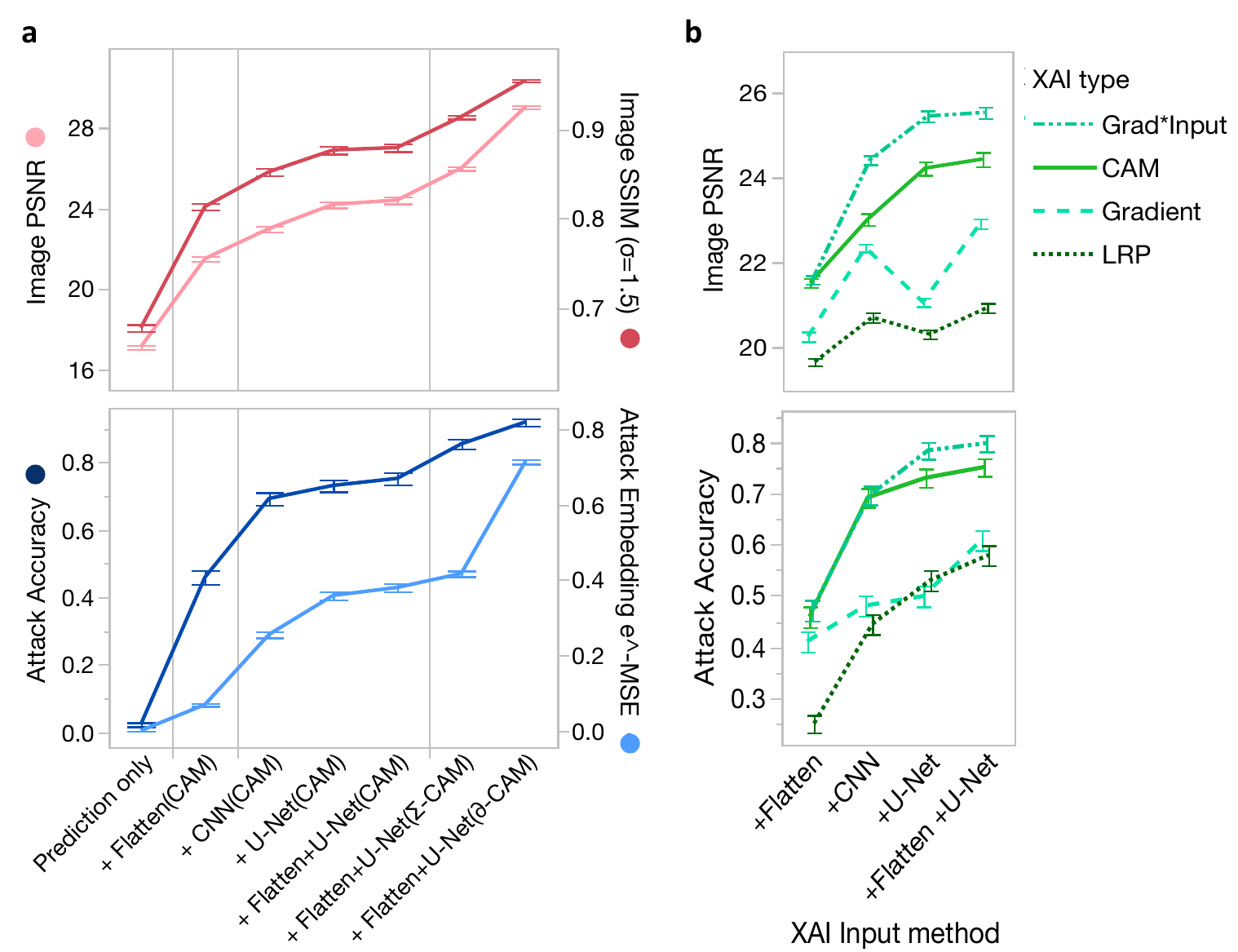}
    \vspace{-0.5cm}
    \caption{
    Inversion attack performance for different XAI input methods and XAI types of iCV-MEFED \cite{loob2017dominant} (Emotion target task). % (Complete version of Figure \ref{fig:xaiInputs}b).
    PSNR replacing $1-\text{MSE}$.
    Error bars indicate 90\% confidence interval.
    }
    \label{fig:xaiInputs_sup}
\end{figure}

We calculated the peak signal-to-noise ratio (PSNR) as another popular image similarity metric, where $\text{PSNR}=log_{10}(\text{MAX}^2/\text{MSE})$ and $\text{MAX}$ refers to the dynamic range of the image pixel (255 for 8-bit greyscale images). Supplementary Figure \ref{fig:xaiInputs_sup}a shows the inversion performance with PSNR replacing $1-\text{MSE}$ in Figure \ref{fig:xaiInputs}a in the main paper.

In addition to Pixelwise Similarity $1-\text{MSE}_x$ and Attack Embedding Similarity $e^{-\text{MSE}_s}$ reported in Figure \ref{fig:xaiInputs}b of the main paper, we report Image PSNR and Attack Accuracy as alternative metrics for inversion attack performance (Supplementary Figure \ref{fig:xaiInputs_sup}b).

\begin{figure*}[ht]
    \centering
    \hspace*{-0.5cm}
    \includegraphics[width=16.4cm]{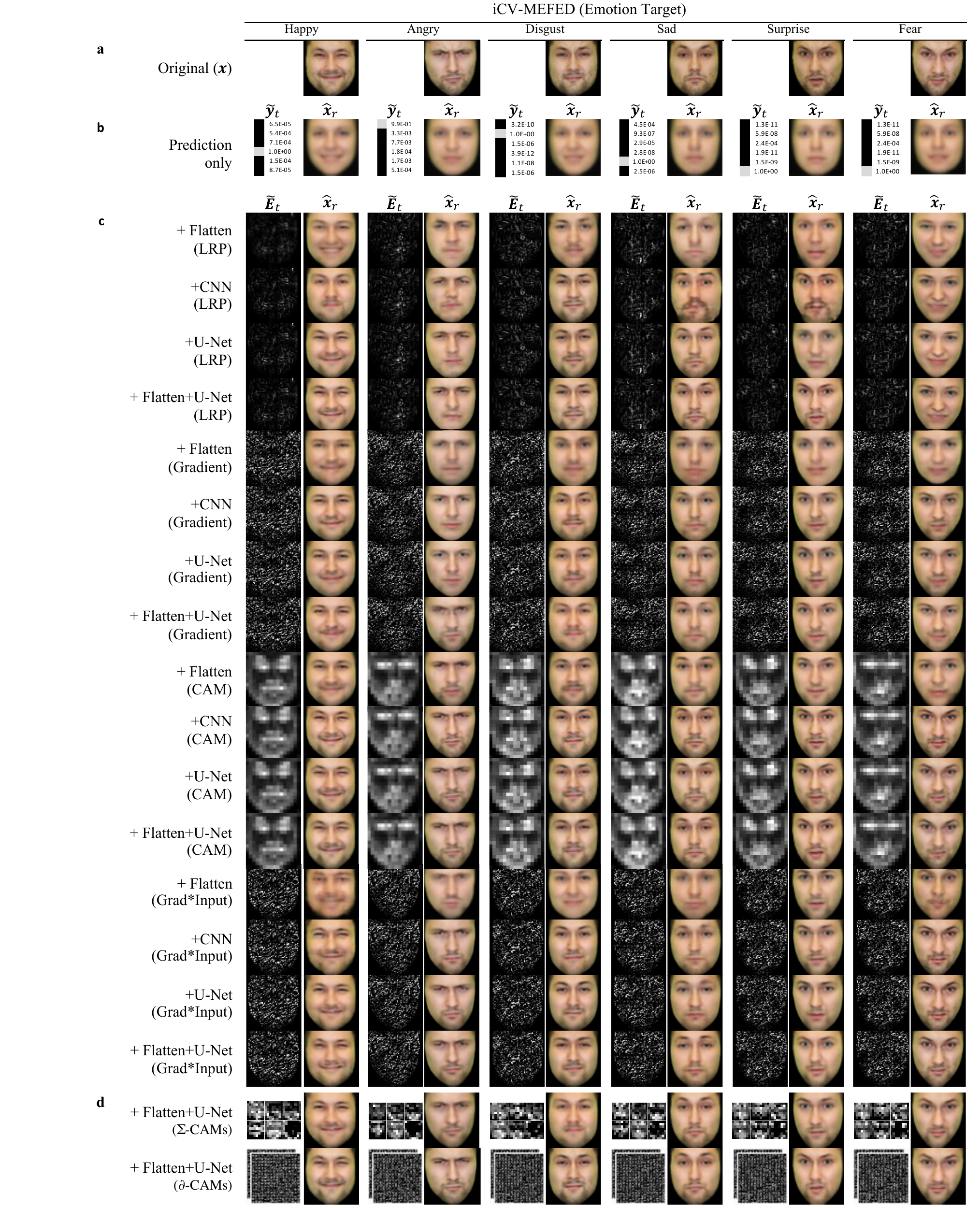}
    \caption{
    Demonstration of image reconstruction from XAI-aware inversion attack with emotion prediction as the target task, and face reconstruction as the attack task. Six emotions of a single identity come from the iCV-MEFED dataset \cite{loob2017dominant}. Reconstructed images are shown with corresponding information (ie. target prediction $\tilde{\bm{y}}_t$, explanations $\tilde{\bm{E}}_t$ as Gradients \cite{simonyan2013deep}, Grad-CAM \cite{selvaraju2017grad} or $\text{Gradient} \odot \text{Input}$ \cite{shrikumar2016not} saliency maps). Towards original images (a), reconstructions from Prediction only (b) are poor and similar across different faces, and are significantly improved when exploiting single (c) and multiple (d) explanations.
    Fredrikson baseline not used, since it is a class-based inversion and will only have the same reconstructed image for each emotion class, regardless of face identity or instance.
    }
    \label{fig:emo_target}
    \vspace{-0.8cm}
\end{figure*}

Supplementary Figure \ref{fig:emo_target} demonstrates more instances with explanations and image reconstructions across all XAI Input Methods.

\subsection{Attacking non-explainable target models}
% {\color{red}
% We report the performance comparison with another baseline raised by Fredrikson et al. \cite{fredrikson2015model}, to show the superiority of our method. 

% Note that Fredrikson is a class-based inversion technique (i.e. reconstructing image w.r.t. a class label) while the other approaches are instance-based (i.e. reconstructing image w.r.t. an image instance).

% Supplementary Figures \ref{fig:icv_id}, \ref{fig:celeba} and \ref{fig:mnist} demonstrate more instances with explanations and image reconstructions across non-explainable and explainable target models.
% }

% \newpage
% \and

\begin{figure}[h]
    \centering
    \includegraphics[width=8.2cm]{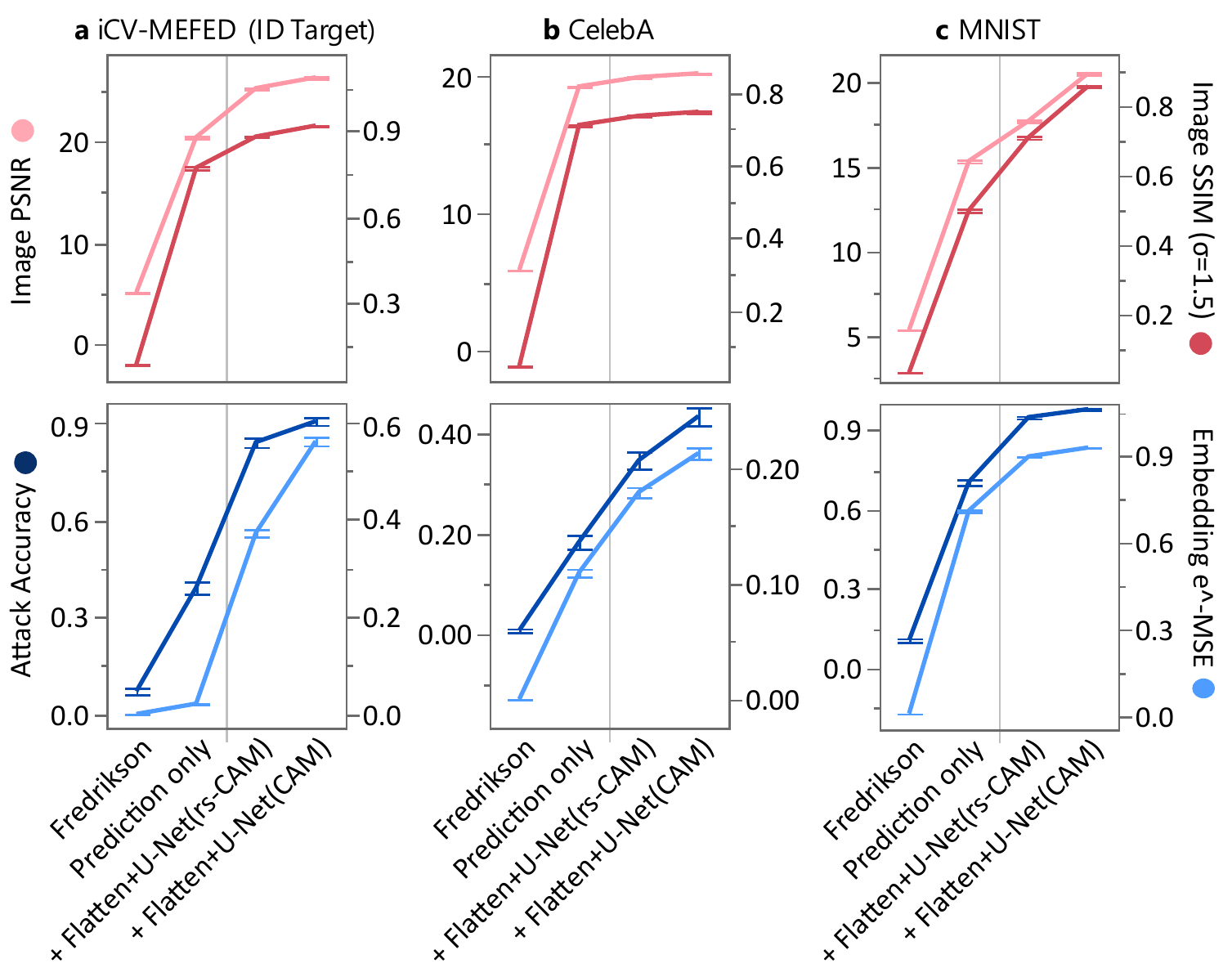}
    \vspace{-.1cm}
    \caption{
    Inversion attack performance across different datasets showing increased privacy risk when exploiting target explanations (CAM) and with attention transfer.
    % Performance with baseline method by Fredrikson et al. \cite{fredrikson2015model} is significantly poorer than training inversion attack model with target model prediction only.
    Two non-XAI-aware baselines Fredrikson \cite{fredrikson2015model} and Prediction only \cite{yang2019neural} are significantly poorer than XAI-aware inversions.
    % Even without target explanations, inversion performance with reconstructed, surrogate explanations (rs-CAM) was similar to exploiting target explanations.
    Error bars indicate 90\% confidence interval.
    % Inversion attack performance across different datasets when introducing Fredrikson et al. \cite{fredrikson2015model} for comparison (supplementary for Figure. \ref{fig:xaiAttention}). 
    }
    \label{fig:xaiAttention_supplementary}
\end{figure}

\begin{figure}[h]
    \centering
    \includegraphics[width=5.8cm]{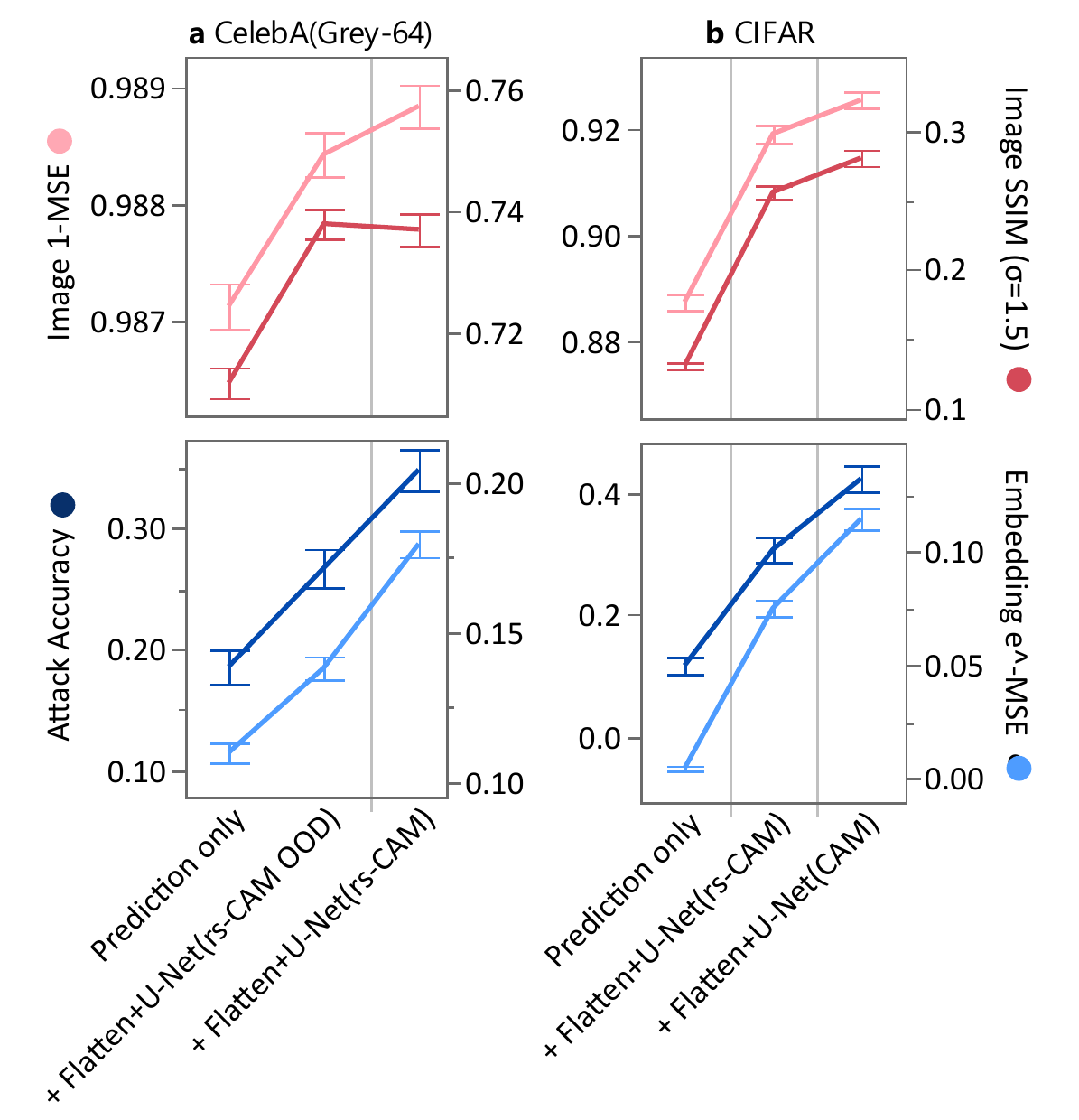}. % resizing due to only 2 columns 
    \vspace{-.1cm}
    \caption{
    % \textcolor{green}{Inversion attack performance on CelebA showing increased privacy risk when exploiting OOD and IID and attack dataset with attention transfer;OOD training (rs-CAM(OOD)) can still perform improved inversion attack, albeit weaker.  Inversion attack performance on CIFAR showing increased privacy risk when exploiting target explanations (CAM) and with attention transfer. 
    % Error bars indicate 90\% confidence interval.}
    Inversion attack performance for attack model trained on OOD data (a) and for CIFAR data (b) 
    showing increased privacy risk when exploiting target explanations (CAM) and with attention transfer.
    Error bars indicate 90\% confidence interval.
    }
    \label{fig:ood}
\end{figure}

Inversion performance improved in the order: Fredrikson $<$ Prediction only $<$ rs-CAM $<$ CAM (Supplementary Figure \ref{fig:xaiAttention_supplementary}). 
Fredrikson has poorest performance because it performs class inversions (i.e., only one reconstructed image per class), while the other approaches are instance inversions (i.e., reconstruct differently per instance). 
For ecological validity regarding privacy, we evaluated the inversion attacks on out-of-distribution (OOD) data (FaceScrub~cite{ng2014data}) and a real-world photo dataset (CIFAR-10 \cite{krizhevsky2009learning}). Supplementary Fig. \ref{fig:ood}a shows that attack models trained with OOD data still perform improved inversion attack compared to prediction only, albeit weaker than models trained with independent and identically distributed (IID) data~\cite{liu2018large}. Supplementary Fig. \ref{fig:ood}b shows similar inversion attack performance with CIFAR compared to iCV-MEFED, CelebA, and MNIST.

Supplementary Figures 
% \ref{fig:icv_id}, \ref{fig:celeba} and \ref{fig:mnist} 
\ref{fig:icv_id}-\ref{fig:cifar_sample}
demonstrate explanations and image reconstructions more demonstration instances for baseline and XAI-aware, and XAI Input methods for non-explainable and explainable target models.

% \begin{figure}[h]
%     \centering
%     % \includegraphics[width=6.9cm]{figures/xaiInputs.pdf}
%     % \includegraphics[width=11.7cm]{figures_supp/xairelation_supp.pdf}
%     \includegraphics[width=8.2cm]{figures_rgb/appendix_xai_input.pdf}
%     \vspace{-.2cm}
%     \caption{
%     % Inversion attack performance for different XAI input methods and XAI types
%     % of iCV-MEFED (Emotion target task)
%     % increases for 
%     % a) more spatially-aware architectures and with multi-explanations, and 
%     % b) explanation types that include sharper details of the input image.
%     % Error bars indicate 90\% confidence interval.
%     Inversion attack performance for different XAI input methods and XAI types of iCV-MEFED \cite{loob2017dominant} (Emotion target task). % (Complete version of Figure \ref{fig:xaiInputs}b).
%     Error bars indicate 90\% confidence interval.
%     }
%     \label{fig:xairelation}
%     \vspace{-.4cm}
% \end{figure}

\pagebreak

\renewcommand{\thefigure}{\arabic{figure} (Cont.)}
\addtocounter{figure}{-1}
\begin{figure*}[ht]
    \centering
    % \ContinuedFloat
    % \captionsetup{list=off,format=cont}
    \hspace*{-0.5cm}
    \includegraphics[width=10.4cm]{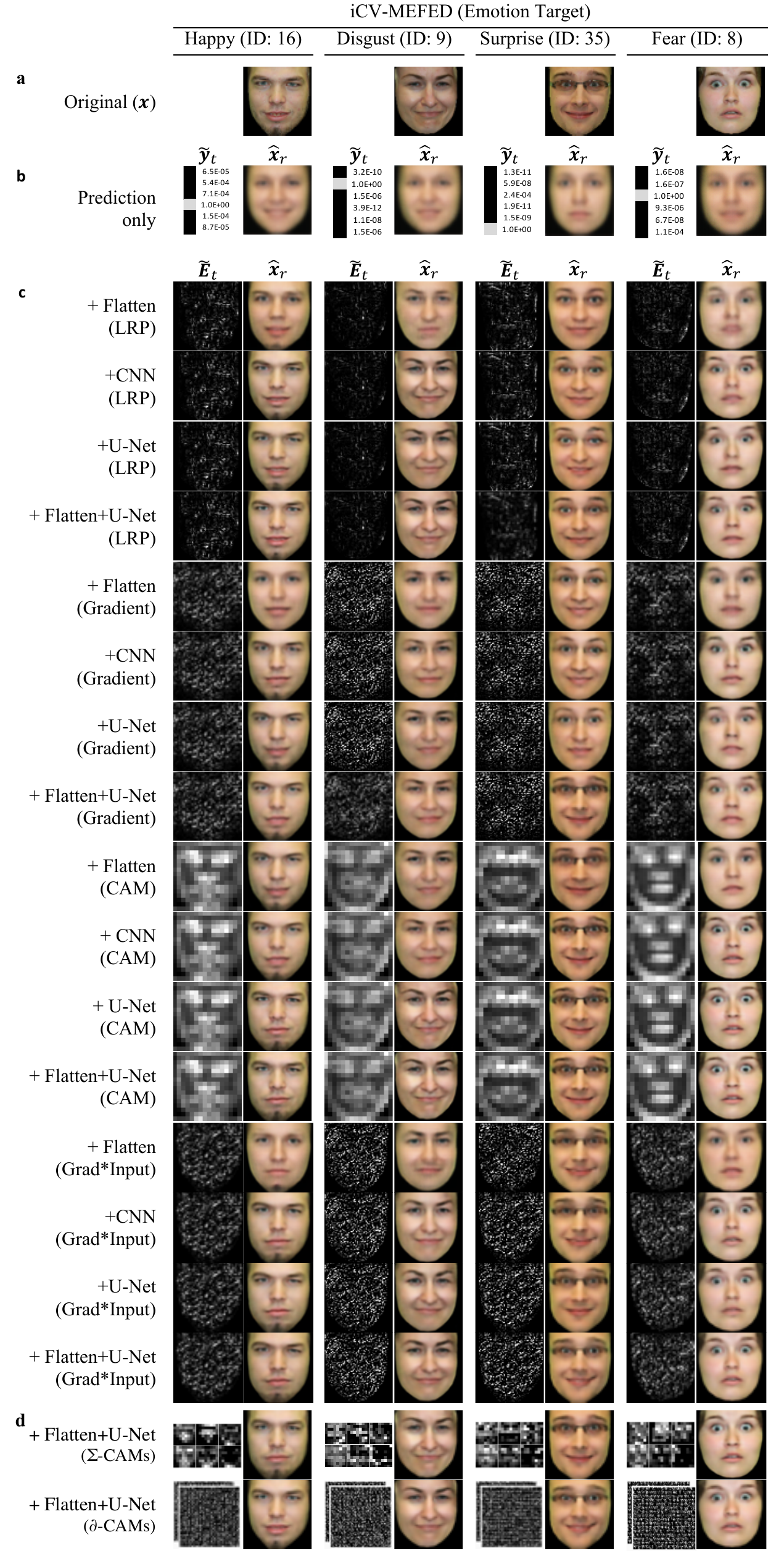}
    \caption{
    Demonstration of image reconstruction from XAI-aware inversion attack with emotion prediction as the target task, and face reconstruction as the attack task. Four emotions of different identities come from the iCV-MEFED dataset \cite{loob2017dominant}.
    }
    \label{fig:emo_continue}
    \vspace{-0.8cm}
\end{figure*}
\renewcommand{\thefigure}{\arabic{figure}}

% \pagebreak
\clearpage
\begin{figure*}[h]
    \centering
    \includegraphics[width=17.4cm]{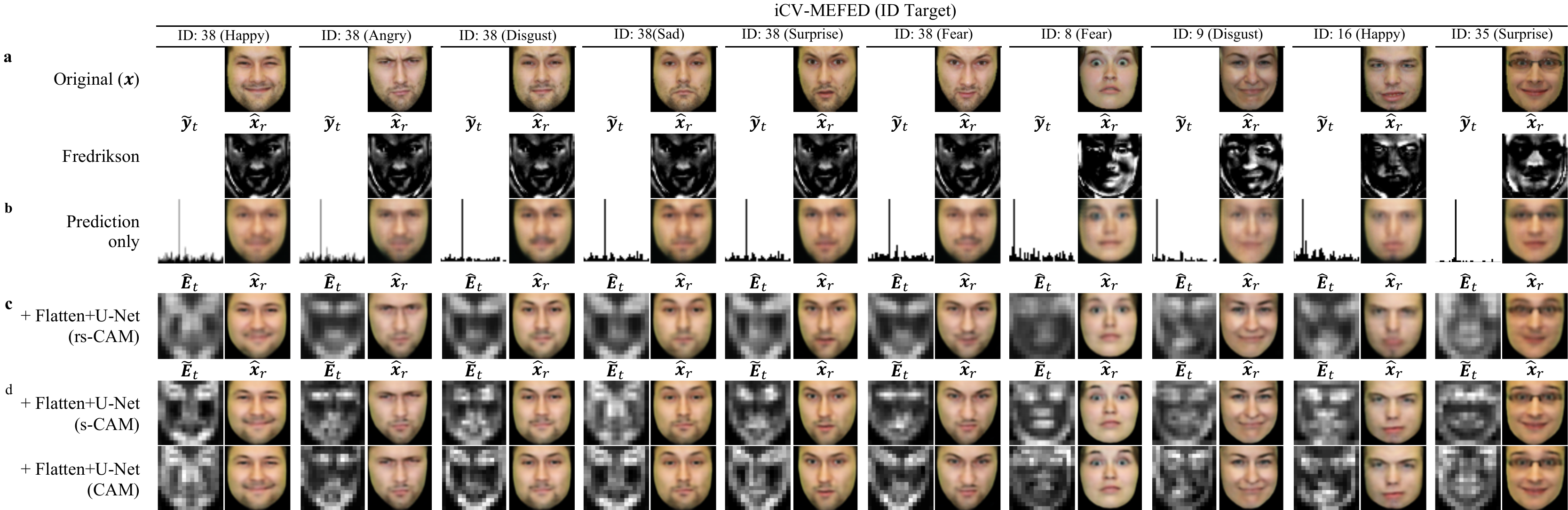}
    % \vspace{-0.7cm}
    \caption{
    Demonstration of image reconstruction with baseline and XAI-aware inversion attack models for iCV-MEFED \cite{loob2017dominant} with identification as target and attack tasks. 
    % Towards original images (a), reconstructions from Fredrikson model inversion \cite{fredrikson2015model} and Prediction only (b) are poor and similar across different faces, and is significantly improved when exploiting reconstructed surrogate CAM(c), surrogate CAM and target CAM(d) explanations.
    }
    \vspace{-0.0cm}
    \label{fig:icv_id}
\end{figure*}

\begin{figure*}[h]
    \centering
    \includegraphics[width=17.4cm]{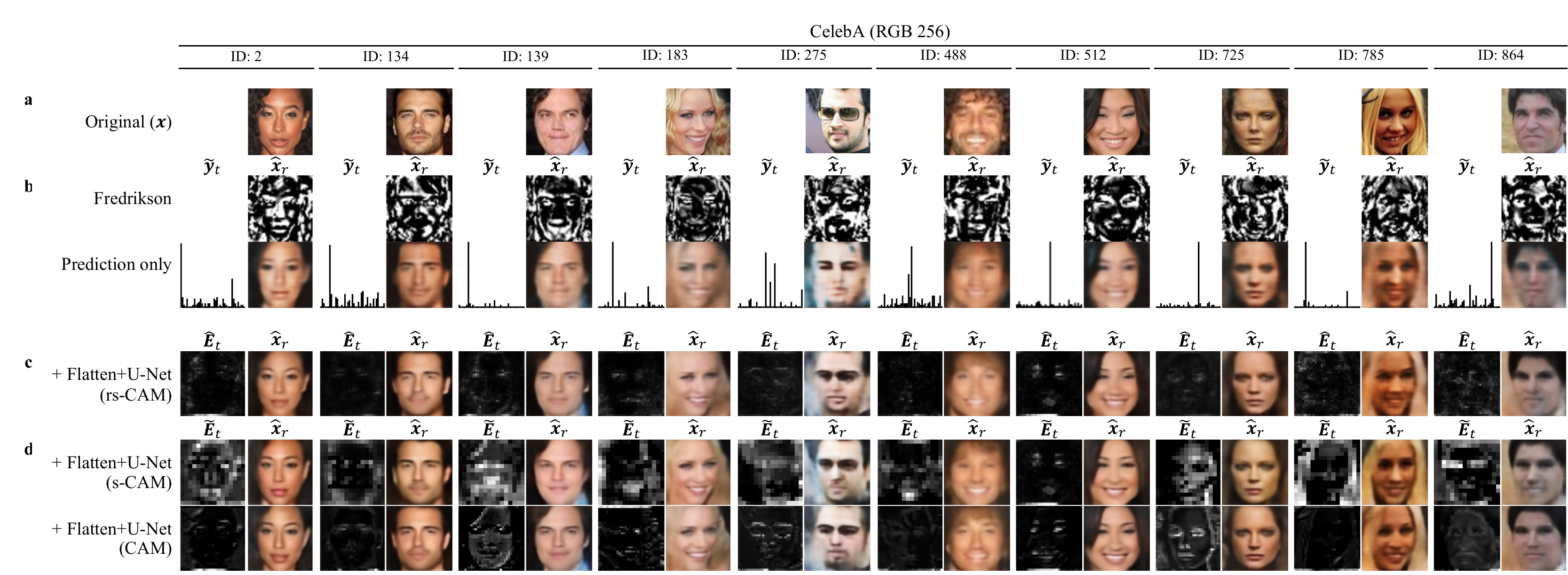}
    % \vspace{-0.7cm}
    \caption{
    Demonstration of image reconstruction with baseline and XAI-aware inversion attack models for CelebA \cite{liu2018large} with identification as target and attack tasks.% Same format as Supplementary Figure \ref{fig:icv_id}.
    % CelebA faces: Madison Pettis (?), Ann Curry, Larry David
    }
    \vspace{-0.0cm}
    \label{fig:celeba}
\end{figure*}

\begin{figure*}[h]
    \centering
    \includegraphics[width=17.4cm]{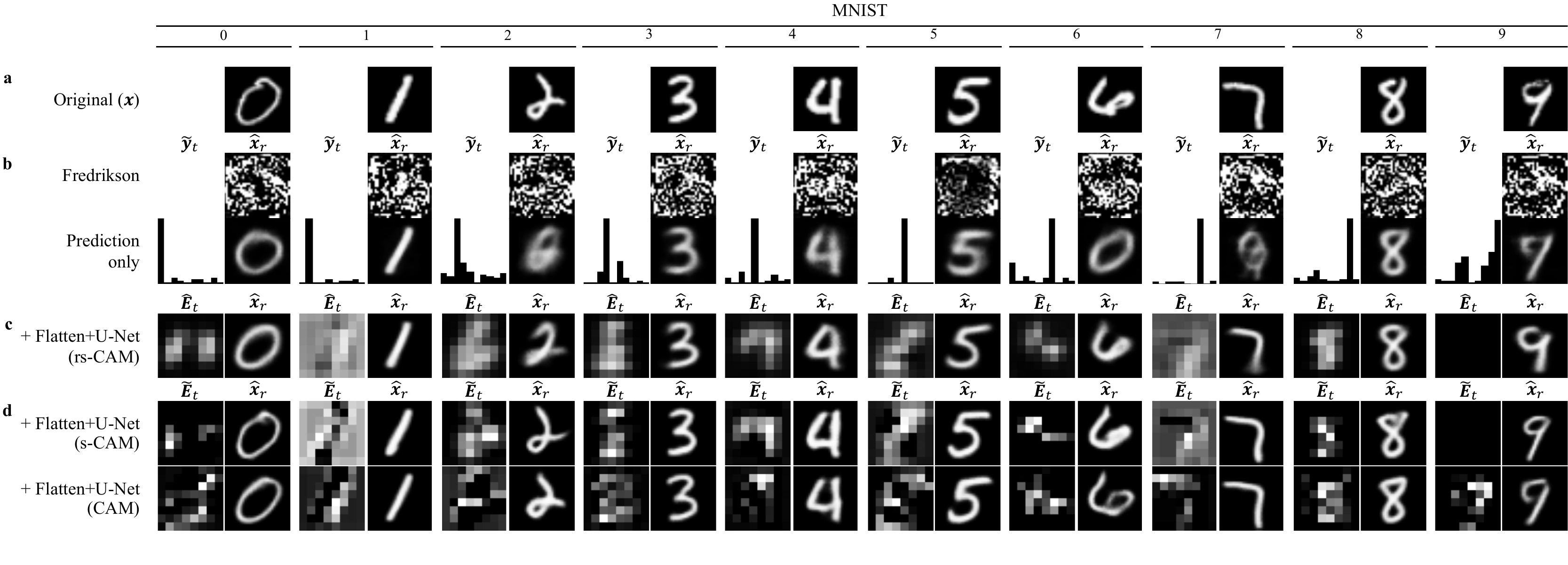}
    \vspace{-0.7cm}
    \caption{
    Demonstration of image reconstruction with baseline and XAI-aware inversion attack models for MNIST \cite{lecun1998gradient} with handwriting digit recognition as target and attack tasks.% Same format as Supplementary Figure \ref{fig:icv_id}.
    % CelebA faces: Madison Pettis (?), Ann Curry, Larry David
    }
    \vspace{-0.0cm}
    \label{fig:mnist}
\end{figure*}

\begin{figure*}[h]
    \centering
    \includegraphics[width=17.4cm]{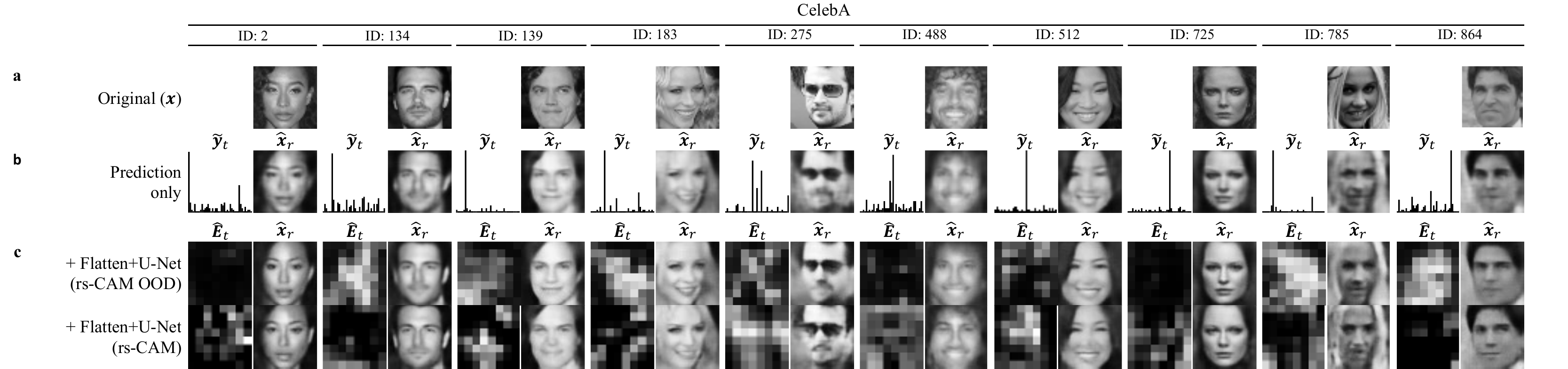}
    % \vspace{-0.7cm}
    \caption{
    Demonstration of image reconstruction with baseline and XAI-inversion attack models trained on OOD data (FaceScrub~\cite{ng2014data}) dataset to attack a target model (trained on (CelebA~\cite{liu2018large})
    .% Same format as Supplementary Figure \ref{fig:icv_id}.
    % CelebA faces: Madison Pettis (?), Ann Curry, Larry David
    }
    \vspace{-0.0cm}
    \label{fig:celeba_ood}
\end{figure*}

\begin{figure*}[h]
    \centering
    \includegraphics[width=17.4cm]{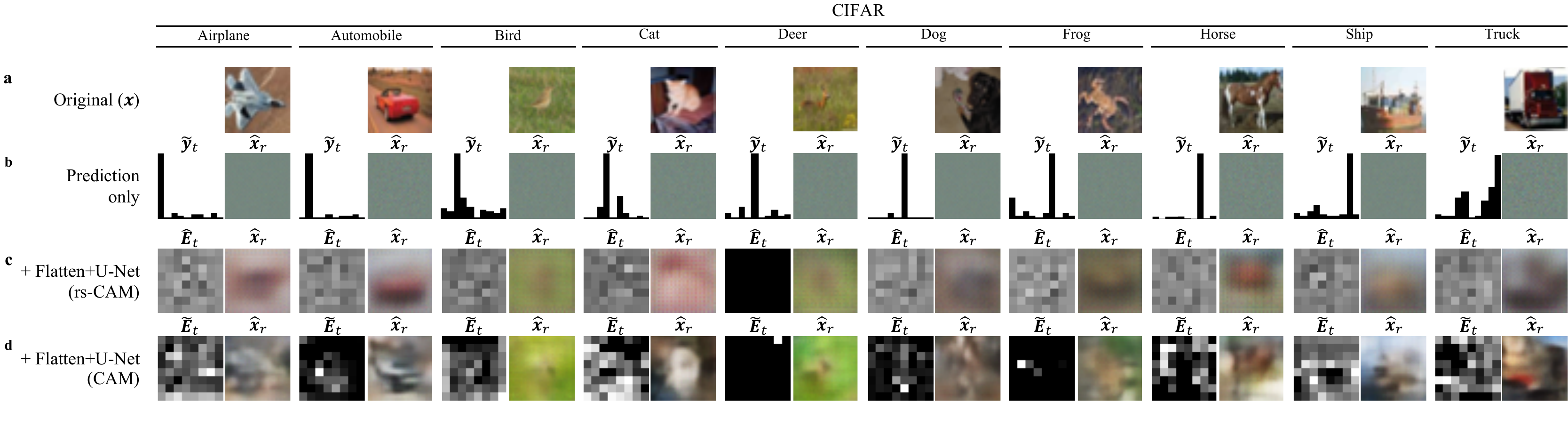}
    \vspace{-0.7cm}
    \caption{
    Demonstration of image reconstruction with baseline and XAI-aware inversion attack models for CIFAR-10 with recognition as target and attack tasks.% Same format as Supplementary Figure \ref{fig:icv_id}.
    % CelebA faces: Madison Pettis (?), Ann Curry, Larry David
    }
    \vspace{-0.0cm}
    \label{fig:cifar_sample}
\end{figure*}

\end{document}